\documentclass[journal]{IEEEtran}

\usepackage{amssymb}
\usepackage{amsfonts}
\usepackage{amsmath}
\usepackage{graphicx}
\usepackage{graphics}
\usepackage{cite}
\usepackage{color}
\usepackage{multirow}
\usepackage{boxedminipage}

\input{mysymbol.sty}
\input{mysymboltensor.sty}
\def\Tr{\mathsf{T}}

\usepackage{accents}

\usepackage{amsthm}
\newtheoremstyle{sltheorem}
{}                
{}                
{\slshape}        
{}                
{\bfseries}       
{.}               
{ }               
{}                
\theoremstyle{sltheorem}

\newtheorem{corollary}{Corollary}
\newtheorem{lemma}{Lemma}
\newtheorem{proposition}{Proposition}

\renewcommand{\baselinestretch}{.97}

\definecolor{mygreen}{rgb}{0.10,0.50,0.10}

\begin{document}

\title{Tensor and Matrix Low-Rank Value-Function Approximation in Reinforcement Learning}

\author{Sergio~Rozada,~\IEEEmembership{Student Member,~IEEE,} Santiago~Paternain, ~\IEEEmembership{Member,~IEEE,}
        and~Antonio~G.~Marques,~\IEEEmembership{Senior Member,~IEEE}\thanks{S. Rozada and A. G. Marques are with the Department
of Signal Theory and Comms., King Juan Carlos University, Madrid, Spain, s.rozada.2019@alumnos.urjc.es, antonio.garcia.marques@urjc.es. S. Paternain is with Department of Electr., Computer and Systems Eng. Rensselaer Polytechnic Institute, Troy, USA, paters@rpi.edu.}%
\thanks{Work partially funded by the Spanish NSF (MCIN/AEI/10.13039/ 501100011033) Grants PID2019-105032GB-I00, TED2021-130347B-I00, and PID2022-136887NB-I00, the ``European Union NextGenerationEU/PRTR'', and the Autonomous Community of Madrid (ELLIS Madrid Unit).}}

\maketitle
\vspace{-0.8cm}
\begin{abstract}
Value function (VF) approximation is a central problem in reinforcement learning (RL). Classical non-parametric VF estimation suffers from the curse of dimensionality. As a result, parsimonious parametric models have been adopted to approximate VFs in high-dimensional spaces, with most efforts being focused on linear and neural network-based approaches. Differently, this paper puts forth a \emph{parsimonious non-parametric} approach, where we use \emph{stochastic low-rank algorithms} to estimate the VF matrix in an online and model-free fashion. Furthermore, as VFs tend to be multi-dimensional, we propose replacing the classical VF matrix representation with a tensor (multi-way array) representation, and then using the PARAFAC decomposition to design an online model-free tensor low-rank algorithm. Different versions of the algorithms are proposed, their complexity is analyzed, and their performance is assessed numerically using standardized RL environments. 
\end{abstract}


\section{Introduction}

Ours is an increasingly complex and interconnected world, where every day a myriad of digital devices generate continuous flows of information. In order to navigate such heterogeneous landscapes, computer systems need to become intelligent and autonomous to make their own decisions without human supervision. In this context, reinforcement learning (RL) has recently been profiled as a powerful tool to embed human-like intelligence into software agents. At a high level, RL is a learning framework that tries to mimic how humans learn to interact with the environment by trial-and-error \cite{sutton2018reinforcement, bertsekas2019reinforcement}. As in dynamic programming (DP), RL aims to solve a sequential optimization problem where the information comes in the form of a numerical signal (cost or reward) \cite{bertsekas2000dynamic}. However, RL does not assume any prior knowledge (model) of the world and learns just by sampling trajectories from the environment, typically modeled as a Markov decision process (MDP) \cite{sutton2018reinforcement}. RL has recently achieved some important milestones in the Artificial Intelligence (AI) community, with notable examples including AlphaGo, a computer software that beat the world champion of the game Go \cite{silver2016mastering} \cite{silver2017mastering}, or, more recently, chatbots such as chatGPT~\cite{brown2020language}. 

In a nutshell, the RL problem comes down to learning a set of actions to take, termed ``policy'', for every instance of the world, which is modeled as a state that varies over time \cite{sutton2018reinforcement} \cite{bertsekas2019reinforcement}. Value functions (VFs) are a core concept of RL. VFs synthesize and quantify the potential reward that an agent can obtain from every state of the environment. Traditionally, non-parametric models of the VF have been a popular approach to solving RL, with the most celebrated example being $Q$-learning \cite{watkins1992q}. However, non-parametric methods are algorithmically challenging as the size of the environment grows. To overcome this problem, parametric models have been introduced to approximate VFs\footnote{Parametric methods are also heavily used to implement policy-based approaches, which are of paramount importance in DP and RL. While the focus of this paper is on VF-based RL approaches, readers interested in policy-based approaches are referred to \cite{dong2020deep, sutton2018reinforcement}} \cite{bertsekas1996neuro}. A simple but effective solution is to consider a linear approach under which the value for each state is modeled as a weighted sum of the features associated with that state \cite{melo2007q}. The parameters  to be estimated are the weights, and the key issue is how to define the set of features, with options including leveraging prior knowledge of the problem structure, using clustering data-based approaches, or exploiting spectral components of the Markovian transition matrices, to name a few \cite{behzadian2019fast, behzadian2018low}. Thanks to the rise of deep learning (DL), parametric VF methods using different neural network (NN) architectures have emerged as a promising alternative for RL \cite{mnih2013playing} \cite{mnih2015human}. The parameters correspond to the linear weights inside each of the layers, whose values are tuned using a backpropagation algorithm. While powerful, one of the main disadvantages of NN-based approximators is that they are not easy to interpret. In contrast, classical VF estimation involve tabular methods with direct interpretation. VFs are typically represented as matrices, with rows indexing states, columns indexing actions, and the entries of the matrix representing the long-term reward associated with a particular state-action pair. 
If a VF is properly estimated, the optimal action for a given state can be chosen by extracting the row corresponding to the state and then finding the column with the largest value. Surprisingly, while the matrix VF representation is pervasive, most RL and VF approximation methods do not take advantage of the underlying matrix structure. For that reason, this work puts forth low-rank VF models and algorithms that give rise to efficient \emph{online} and \emph{model-free} RL schemes. Relative to traditional non-parametric methods, the consideration of low-rank matrices accelerates convergence, enhancing both computational and sample complexity. Moreover, since most practical scenarios deal with high-dimensional state-action spaces, we take our approach one step further, postulating the reshaping of the VF matrix as a tensor, and designing \emph{tensor-based low-rank RL schemes} that can further reduce computational and sample complexity.

Matrix low-rank regularization has been successfully employed in the context of low-rank optimization and matrix completion \cite{eckart1936approximation, markovsky2012low,udell2016generalized, mardani2013decentralized}, but its use in DP and RL has been limited. Some of the most relevant examples are briefly discussed next. Matrix-factorization schemes have been utilized to approximate the transition matrix of an MDP \cite{barreto2016incremental, jiang2017contextual}, which are subsequently applied to obtain the associated VF and optimal policy. Interestingly, when the transition matrix is low rank, model-based approaches that leverage this property have been shown to have low sample complexity \cite{agarwal2020flambe, uehara2021representation}. Other works proposed obtaining first the VF of an MDP, and then approximating it using a low-rank plus sparse decomposition \cite{ong2015value}. On a similar note, a rank-1 approximation of the VF has been proposed in the context of energy storage \cite{cheng2016co, cheng2017low}. More recently, \cite{yang2019harnessing, shah2020sample} proposed to approximate the VF via low-rank optimization in a \emph{model-based} and \emph{offline} setup. Finally, \cite{sam2023overcoming} dealt with a setup similar to those in \cite{yang2019harnessing, shah2020sample}, but it focused on the sample complexity of low-rank VF methods, given a generative model of the MDP.

In parallel, the use of tensor methods for generic machine-learning applications has been on the rise during the last years \cite{kolda2009tensor,sidiropoulos2017tensor,vasilache2018tensor}. Recent advances in low-rank tensor decomposition (especially in the context of efficient optimization) and the ubiquity of multi-dimensional datasets have been key elements fostering that success. As high-dimensional state-action spaces are fairly common in RL, tensors are a natural structure to model VF. Moreover, tensor low-rank regularization techniques are efficient in terms of the number of parameters\cite{sidiropoulos2017tensor}, which helps to alleviate the curse of dimensionality problem in RL. However, despite their versatility and ease of interpretation, the use of tensors in the context of RL is limited. Tensor-decomposition algorithms have been proposed to estimate the dynamics of MDPs in \emph{model-based} RL \cite{mahajan2021tesseract, van2021model, azizzadenesheli2016reinforcement_a, azizzadenesheli2016reinforcement_b}. Recent work also showed the potential of tensor-based models in the so-called ``learning-from-experts'' problems \cite{guo2019hybrid}. In the context of network routing, \cite{tsai2021tensor} proposed to estimate a tensor-based VF model using $Q$-learning, but they introduce an interpolation step every fixed-number of episodes that solves a \emph{synchronous} and \emph{offline} tensor-decomposition problem. 

Motivated by the previous context, this paper leverages low-rank matrix and tensor decomposition to develop new RL models and algorithms for the non-parametric estimation of the VF. Our particular contributions include the following.
\begin{itemize}
\item A formulation of the VF approximation problem with tensor-based models.
\item \emph{Stochastic}, \emph{online}, \emph{model-free} algorithms that impose low rank in either the VF matrix or the VF tensor.
\item Convergence guarantees for a \emph{synchronous} model-based algorithm that imposes low rank in the VF matrix as well as for its generalization to the tensor case.
\item Numerical experiments illustrating the practical value of the proposed schemes.
\end{itemize}

\section{Preliminaries}

The goal of this section is to introduce notation, the fundamentals of RL (including the formal definition of the VF and its matrix form), and a brief discussion on low-rank methods. 
\subsection{Reinforcement Learning}

RL deals with a framework where agents interact sequentially with an environment in a closed-loop setup \cite{sutton2018reinforcement} \cite{bertsekas2019reinforcement}. The environment is described by a set of (discrete) states $\ccalS$, where one or more agents take actions from a set of (discrete) actions $\ccalA$. We refer to the cardinality of the state space $\ccalS$ and the action space $\ccalA$ as $C_\ccalS$ and $C_\ccalA$ respectively. With $t=1,...,T$ representing the time index, at time $t$ the agent takes a particular action $a_t$ at a given state $s_t$ and receives a reward $r_t$, that quantifies the instantaneous value of that particular state-action pair. Then, the agent reaches a new state $s_{t+1}$. The transition between states is assumed to be Markovian and governed by the transition probability function 
$\ccalP^a_{s s^{'}} = \Pr{s_{t+1}=s'|s_t\!=\!s, a_t\!=\!a}$. The dependence of $r_t$ on $s_t$ and $a_t$ can be stochastic. Its expectation,  $\ccalR^a_{s s^{'}} = \mathbb{E}[r_{t+1}=r|s_t\!=\!s, a_t\!=\!a, s_{t+1}\!=\!s']$ is termed the reward function.  A fundamental aspect of RL is that making a decision on the value of $a_t$ not only impacts $r_t$, but also subsequent $s_{t'}$ for $t' > t$. Since $r_{t'}$ for $t' > t$ depends on $s_{t'}$, this readily implies that the optimization (i.e., the selection of the optimal action at time $t$) is coupled across time. Recasting the optimization problem as an MDP is a common approach to deal with the time dependence of the optimization and the stochasticity of the rewards. RL can be then understood as a stochastic approach for solving the MDP described by $\langle \ccalS, \ccalA, \ccalP_{ss'}^a, \ccalR_{ss'}^a \rangle$.

Mathematically, the goal of RL is to find a policy $\pi : \ccalS \mapsto \ccalA$ that maps states to actions optimally. Given a particular policy $\pi$, let us define the state-action value function (SA-VF) $Q^\pi: \ccalS \times \ccalA \mapsto \reals$ as the expected aggregated reward of each state-action pair $\mathbb{E}^{\pi}[\sum_{\tau=t}^{\infty}\gamma^{\tau-t}r_\tau|s_t, a_t]$, where $\gamma\in(0,1)$ is a discount factor that places more focus on near-future values. The $Q^\pi$ values of each state-action pair can be arranged in the form of a matrix $\bbQ^\pi \in \reals^{C_\ccalS \times C_\ccalA}$, usually referred to as the $Q$-matrix. Among all possible SA-VFs, there exists an optimal $Q^*(s,a) = \max_\pi Q^\pi(s,a) ~\forall ~(s, a) \in \ccalS \times \ccalA$. The optimal policy $\pi^*$ is then simply given as $\pi^*(s) = \argmax_a Q^*(s, a) ~\forall s \in \ccalS$. It is then clear that the SA-VF plays a central role in RL for estimating the optimal policy of the MDP.

One particular feature of RL is that the statistical model that relates the states and the actions is unknown (model-free). The learning takes place on the fly using trajectories sampled from the MDP to estimate the optimal policy $\pi^* $. This can be accomplished either by directly optimizing a parametric model of $\pi$ or via learning the SA-VF. This results in two families of algorithms: policy gradient RL and value-based RL \cite{sutton2018reinforcement}, the latter of which is the focus of this manuscript as mentioned in the introduction.
%
%
Value-based RL is founded in non-parametric methods that try to estimate $\bbQ^\star$, the optimal $Q$-matrix of the MDP, with the $Q$-learning being the original approach. While works remarkably well when the cardinalities $C_\ccalS$ and $C_\ccalA$ are small, in more complex environments where the state and actions involve multiple dimensions, the size of the $Q$-matrix grows exponentially, hindering the adoption of non-parametric $Q$-learning. Parsimonious parametric alternatives, such as linear models and NNs, that approximate the VF via the estimation of a finite set of parameters were put forth to alleviate the high-dimensionality problem. 
More specifically, linear models postulate schemes of the form $[\hbQ]_{s,a}\!=\!Q_{\beta}(s, a)\!=\!\bbphi(s, a)^\top \bbbeta$, where $\bbphi(s, a)\in \reals^P$ is the set of $P$ features that defines a particular state-action pair and $\bbbeta\in \reals^P$ are the parameters/weights to be estimated. Ideally, when defining the features these VF methods try to exploit the structure of the $Q$-matrix $\bbQ$ to leverage compact representations that scale to larger state-action spaces. On the other hand, NNs are used to leverage more complex (non-linear) mappings to estimate $\hbQ$, so that $[\hbQ]_{s,a}\!=\!Q_{\omega}(s, a)\!=\!\ccalT(\phi (s, a), \bbomega)$, where $\ccalT(\cdot)$ is the NN mapping and $\bbomega$ the parameters of the NN. 

As explained in greater detail in the ensuing sections, our approach is to develop parsimonious models to estimate the SA-VF by imposing constraints on the rank of the $Q$-matrix. This not only provides a non-parametric alternative to limit the degrees of freedom of the SA-VF, but also preserves some benefits of the aforementioned parametric schemes including interpretability (as in linear models) and the ability to implement non-linear mappings (as in NNs) capable of capturing more complex dependencies between actions and states. 

\subsection{Low-rank and matrix completion}

While the most popular definition of the rank of a matrix is the dimension of its column space, when dealing with data-science applications decomposition-based definitions of the rank are more useful. In particular, if a matrix $\bbX\in \reals^{N \times M}$ has rank $K\!\leq \!\min\{N,M\}$, then it holds that the matrix can be written as $\bbX = \sum_{k=1}^K \bbX_k$ with $\bbX_k$ being a rank-1 matrix, i.e., with $\bbX_k$ being the outer product of a column and a row vector. Indeed, let the singular value decomposition (SVD) of $\bbX$ be given by $\bbX = \bbU \bbSigma \bbV ^ \Tr$, where the columns of $\bbU\in\reals^{N\times K}$ and $\bbV\in\reals^{M\times K}$, which are orthonormal, represent the singular vectors and $\bbSigma\in\reals^{K\times K}$ is a diagonal matrix with nonnegative entries collecting the singular values. Then, a rank-1 decomposition of $\bbX$ can be written as
\begin{eqnarray}\label{E:SVD_Rank1_decomposition}
&\bbX = \sum_{k=1}^{\rank(\bbX)}\sigma_k \bbu_k \bbv_k^ \Tr&
\end{eqnarray}
where $\sigma_k>0$ is the $k$th singular value, and $\bbu_k$ and $\bbv_k$ the $k$th right and left singular vectors, respectively. Clearly, the SVD can then be used to identify the rank of the matrix as well as to decompose it into the sum of $K$ rank-1 matrices, each of them spanning an orthogonal subspace. More importantly, the SVD can be used to find low-rank approximations. Consider the minimization problem $\min_\hbX ||\bbX - \hbX||_F^2$ subject to $\rank(\hbX) \leq K'$ and suppose that $\rank(\bbX)>K'$. Then, the Eckart-Young theorem \cite{eckart1936approximation} states the $\hbX$ that minimizes the approximation error is given by the truncated SVD 
\begin{eqnarray}\label{E:truncatedSVD_closedformRank1}
&\hbX = \mathrm{TSVD}_{K'}(\bbX):=\sum_{k=1}^{K'} \sigma_k (\bbu_k \bbv_k^ \Tr).
\end{eqnarray}
Since the singular values satisfy $\sigma_k\geq \sigma_{k+1}$ and $\|\sigma_k \bbu_k\bbv_k^\Tr\|_F=\sigma_k$, the truncated SVD preserves the rank-1 factors with the largest Frobenious norm. 

It is often the case that matrix $\bbX$ is not perfectly known, due to noisy and incomplete observations. Since the noise is typically full rank, if the matrix $\bbX$ is completely observed, the truncated SVD can be used to remove (filter) the noise in the spectral components (singular vectors) that are orthogonal to the data. The problem is more intricate when the values of the matrix $\bbX$ are observed only in a subset of entries. With $\ccalM $ denoting that subset, the goal in this setup is to solve
\begin{equation}\label{E:matrix_completion_low_rank}
\min_\hbX \!\!\sum_{(i,j)\in \ccalM} ([\bbX]_{i,j} - [\hbX]_{i,j})^2 ~\mathrm{subject~to}~\rank(\hbX) \leq K',
\end{equation} 
which is often referred to as matrix completion. 
In this case, the problem is non-convex and does not have a closed-form solution. However, a number of efficient algorithms, from convex approximations based on the nuclear norm to iterative SVD decompositions, along with theoretical approximation guarantees are available \cite{cai2010singular, candes2010power}. Interestingly, by leveraging the PARAFAC decomposition \cite{hitchcock1927expression}, similar approaches can be used for tensor-denoising and tensor-completion problems.

As detailed next, our approach is to exploit these techniques to estimate a low-rank matrix representation of the SA-VF (Section \ref{S:low_rank_matrix_Q_estimation}) and a low-rank tensor decomposition of a multidimensional SA-VF (Section \ref{S:low_rank_tensor_Q_estimation}).



\section{Matrix low-rank for SA-VF estimation}\label{S:low_rank_matrix_Q_estimation}

This section starts by analyzing some properties of Bellman's equation (BEQ) \cite{bellman1957dynamic} when the $Q$ matrix is forced to be low-rank. We then review the basics of temporal-differences (TD) algorithms in RL and, based on those, present one of the main contributions of the paper: a \emph{matrix low-rank TD algorithm} to solve RL in a \emph{model-free} and \emph{online} fashion.

\subsection{Low-rank in Bellman's equation}

BEQ, which is a necessary condition for optimality associated with any DP, is a cornerstone for any DP and RL schemes, including VF estimation \cite{bertsekas2019reinforcement}. BEQ can be expressed in matrix form. For this purpose, let us define the transition probability matrix $\bbP \in \reals^{C_\ccalS C_\ccalA \times C_\ccalS}$, with each entry modeled by the transition probability function $\ccalP^a_{s s^{'}}$. Similarly, let us define the policy matrix $\bbPi \in \reals^{C_\ccalS \times C_\ccalS C_\ccalA}$, with each entry modeled by the policy $\pi(s',a')=Pr[a'|s']$ if  the row and the column indices depend on $s'$, and $0$ otherwise. Lastly, the reward vector $\bbr \in \reals^{C_\ccalS C_\ccalA}$ has its entries governed by the reward function $\ccalR^a_{s s^{'}}$ averaged over $s'$. Now, BEQ can be expressed as a set of linear equations in matrix form

\begin{equation}\label{E:Belman_Qvectorized}
\bbq^\pi = \bbr + \gamma \bbP \bbPi \bbq^\pi,
\end{equation}

\noindent where $\bbq^\pi \in \reals^{C_\ccalS C_\ccalA}$ is the vectorization of $\bbQ^\pi$. The particular form of BEQ given in \eqref{E:Belman_Qvectorized} is known as \emph{policy evaluation}, since given a fixed policy $\pi$, the true value of $\bbQ^\pi$ can be obtained via \eqref{E:Belman_Qvectorized}. Since \eqref{E:Belman_Qvectorized} can be viewed as a system of linear equations, the value of $\bbq^\pi$ can be calculated by rearranging terms and inverting a matrix $\bbq^\pi = (\bbI - \gamma \bbP \bbPi)^{-1} \bbr$. However, in most practical cases the number of actions and states is large, rendering the matrix inversion challenging. A common approach is then to leverage the fact of BEQ being a fixed-point equation and use the iterative scheme 
\begin{equation}\label{E:Belman_Qvectorized_iterative}
\bbq_{k+1}^\pi = \bbr + \gamma \bbP \bbPi \bbq_k^\pi,      
\end{equation}
with $k$ being an iteration index. Such a scheme is guaranteed to converge and entails a lower computational complexity \cite{santos2004convergence}.
 
Since our interest is in low-rank solutions, a straightforward approach to obtain a low-rank SA-VF is to set the value of the desired rank (say $K$), compute $\bbq^\pi = (\bbI - \gamma \bbP \bbPi)^{-1} \bbr$, and then perform a truncated SVD to keep the top $K$ rank-1 factors. In particular, with: i) $\mathrm{vec}$ denoting the vectorization operator that, given a rectangular $N \times M$ input matrix $\bbX$, generates a column vector stacking the columns of the input matrix

\[
    \mathrm{vec}(\bbX) := \begin{bmatrix}
                            [\bbX]_1\\
                            [\bbX]_2 \\
                            \vdots \\
                            [\bbX]_M
                          \end{bmatrix}
\]

\noindent where $[\bbX]_j$ denotes the $j$th column of $\bbX$ and ii)  with $\mathrm{unvec}$ denoting the inverse operator (with the dimensions clear from the context), we have that $\hbq^\pi = \mathrm{vec} (\mathrm{TSVD}_K (\mathrm{unvec}(\bbq^\pi))$. Alternatively, one can also implement the truncated SVD in \eqref{E:Belman_Qvectorized_iterative} leading to the iteration 
 
 \begin{equation}\label{E:Bellman_low_rank}
 \hbq^\pi_{t+1} = \mathrm{vec} (\mathrm{TSVD}_K (\mathrm{unvec}(\bbr + \gamma \bbP \bbPi \hbq^\pi_{t})).
 \end{equation} 
 
 This scheme converges to a neighborhood of $\bbq^\pi$, the fixed point of the Bellman operator~\eqref{E:Belman_Qvectorized}. This is the subject of the following proposition, which aligns to standard results that bound the single-step approximation error \cite{bertsekas1996neuro}.

\begin{proposition}
\label{theorem}
Let $K>0$ be a constant denoting the rank of the truncated SVD operator $\mathrm{TSVD}_K(\cdot)$, and let $N$ and $M$ be, respectively, the number of rows and columns of the matrix $\mathrm{unvec}(\bbq^\pi)$. Then, upon the definition of 

\begin{equation}
    \label{eq::b_def}
    B = \max_{t=1,2,\ldots}(\min\{N, M\}-K) \sigma_{K+1}(t),
\end{equation}

\noindent with $\sigma_{K+1}(t)$ being the $K+1$th singular value of matrix $\mathrm{unvec}(\bbr + \gamma \bbP \bbPi \hbq^\pi_{t})$, it holds that 
\begin{equation}
\limsup_{t\to\infty}\left\|\hat{\bbq}_t^\pi - \bbq^\pi \right\|_\infty\leq \frac{B}{1-\gamma}.
\end{equation}
Moreover, the convergence to the neighborhood is linear. 
\end{proposition}

\begin{proof}

Let us start by applying the triangle inequality to upper bound the infinite norm of the difference $\hat{\bbq}^\pi_{t+1}-\bbq^\pi$     
\begin{align}\label{E:Triangle}
    \left\| \hat{\bbq}^\pi_{t+1}-\bbq^\pi \right\|_\infty &\leq \left\|\hat{\bbq}^\pi_{t+1}-(\bbr+\gamma \bbP \boldsymbol{\Pi}\hat{\bbq}^\pi_t )\right\|_\infty \nonumber \\
    &+\left\|\bbr+\gamma \bbP \boldsymbol{\Pi}\hat{\bbq}^\pi_t -\bbq^\pi \right\|_\infty.
\end{align}

\noindent To upper bound the first term in the right hand side of \eqref{E:Triangle}, recall that for a generic matrix $\bbX \in \reals^{N \times M}$ whose (ordered) SVD is given by $\{\sigma_i,\bbu_i,\bbv_i\}_{i=1}^{\min\{N,M\}}$, it holds that
\begin{align}
     &\left\| \mathrm{vec}(\mathrm{TSVD}_K(\bbX)) - \mathrm{vec}(\bbX) \right\|_\infty =\left\| \sum_{i=K+1}^{\min\{N,M\}}\sigma_i(\bbu_i\otimes\bbv_i)\right\|_\infty \nonumber\\
     &\leq \sum_{i=K+1}^{\min\{N,M\}}\sigma_i\left\|\bbu_i\otimes\bbv_i\right\|_\infty  \leq \sum_{i=K+1}^{\min\{N,M\}}\sigma_i\nonumber\\
     &\leq (\min\{N, M\}-K) \sigma_{K+1},
\end{align}

\noindent where $\otimes$ stands for the Kronecker product. Then, for all $t\in \{0\} \cup \mathbb{N} $ it follows that

\begin{align}
\label{eq::b_characterization}
    \| \hat{\bbq}^\pi_{t+1}-&(\bbr+\gamma \bbP \boldsymbol{\Pi}\hat{\bbq}^\pi_t ) \|_\infty \leq \\
    &\max_{t=1,2,\ldots}(\min\{N, M\}-K) \sigma_{K+1}(t)=B. \nonumber
\end{align}

\noindent To upper bound the second term in the right-hand side of \eqref{E:Triangle}, let us replace $\bbq^\pi$ with its definition in \eqref{E:Belman_Qvectorized}
\begin{equation}
    \left\|\bbr-\gamma \bbP \boldsymbol{\Pi}\hat{\bbq}^\pi_t -\bbq^\pi \right\|_\infty = \gamma\left\|\bbP\boldsymbol{\Pi}\left(\hat{\bbq}^\pi_t-\bbq^\pi\right) \right\|_\infty.
\end{equation}
Since $\bbP$ and $\boldsymbol{\Pi}$ are probability matrices, their singular values are bounded by $1$. Hence, the previous expression can be upper-bounded by 
\begin{equation}\label{E:contraction}
    \left\|\bbr-\gamma \bbP \boldsymbol{\Pi}\hat{\bbq}^\pi_t -\bbq^\pi \right\|_\infty \leq \gamma\left\|\hat{\bbq}^\pi_t-\bbq^\pi \right\|_\infty.
\end{equation}
Substituting \eqref{eq::b_characterization} and \eqref{E:contraction} in \eqref{E:Triangle} reduces to
\begin{align}\label{eq:convergence_rate}
    \left\| \hat{\bbq}^\pi_{t+1}-\bbq^\pi \right\|_\infty &\leq B
    +\gamma\left\|\hat{\bbq}^\pi_t -\bbq^\pi \right\|_\infty.
\end{align}
Rewriting the previous expression recursively it follows that 
\begin{align}
    \left\| \hat{\bbq}^\pi_{t+1}-\bbq^\pi \right\|_\infty &\leq B\sum_{s=0}^{t}\gamma^s
    +\gamma^t\left\|\hat{\bbq}^\pi_0 -\bbq^\pi \right\|_\infty.
\end{align}
The result follows from considering the $\limsup$ and using that $\gamma \in(0,1)$. The linear convergence follows directly from \eqref{eq:convergence_rate}.
\end{proof}
The neighborhood to which the iteration \eqref{E:Bellman_low_rank} converges is defined by the error in infinity norm between the application of the Bellman operator and the truncated SVD (see \eqref{E:Bellman_low_rank} and \eqref{eq::b_def}). In particular, the error depends on the $(\min\{N,M\}-K)$ smallest singular values of the matrix $\mathrm{unvec}(\left(\bbr+\gamma \bbP \boldsymbol{\Pi}\hat{\bbq}^\pi_t\right))$ [cf. \eqref{E:SVD_Rank1_decomposition}-\eqref{E:truncatedSVD_closedformRank1}], which are always bounded. 
To prove the result we assume that the bound is independent of the current estimate of $\hat{\bbq}_t^\pi$. Furthermore, this bound need not to hold for all $t\geq 0$, but it suffices that there exists $T>0$ such that for all $t\geq T$ \eqref{eq::b_def} holds. Similar to the previous result, if the policy matrix $\boldsymbol{\Pi}$ has probability one for the action with the higher value, the iteration \eqref{E:Bellman_low_rank} converges to a neighborhood of the optimal $Q$-function. While relevant from a theoretical perspective, the model for the environment is typically not available. Indeed, the full system dynamics are rarely known and their sample-based estimation is unfeasible due to the high dimensionality (states and actions) of the problem. As a result, stochastic model-free methods must be deployed, and that is the subject of the next section. 

\subsection{TD-learning}

TD algorithms are a family of \emph{model-free} RL methods that estimate the SA-VF bootstrapping rewards sampled from the environment and current estimates of the SA-VF \cite{sutton2018reinforcement}. In a nutshell, the problem that TD-learning tries to solve is

\begin{eqnarray} \label{E:Q_estimation_plain}
&\hbQ = \argmin_{\bbQ} \sum_{(s,a)\in \ccalM} (q_s^a - [\bbQ]_{s,a})^2&,
\end{eqnarray}


\noindent where $\ccalM$ is the set of state-action pairs $(s, a)$ sampled from the environment, and $q_s^a$ is the target signal obtained from taking action $a$ in state $s$. The target signal $q_s^a$ should be estimated on the fly using the available reward. Leveraging the form of \eqref{E:Belman_Qvectorized_iterative} and replacing the probabilities with the current estimate, one can define $q_s^a=r_s^a+\gamma [\hbQ]_{s',a'}$, where the tuple $(s',a')$ is the subsequent state-action pair resulting from taking the action $a$ in state $s$.

On the one hand, the action $a'$ is the one that maximizes the current estimate of the SA-VF $[\hbQ]_{s',a'}=\max_a [\hbQ]_{s',a}$. On the other hand, the current estimate of \eqref{E:Q_estimation_plain} is updated  based on a time-indexed sample $(s_t,a_t,r_t,s_{t+1})$. This corresponds to the $Q$-learning update rule:
\begin{align}
[\hbQ^t]_{s_t,a_t} =&  [\hbQ^{t-1}]_{s_t,a_t} + \alpha_t \big( r_t + \gamma \max_a [\hbQ^{t-1}]_{s_{t+1},a}   \nonumber\\
 -& [\hbQ^{t-1}]_{s_t,a_t}\big),\label{E:Q_learning_estimation}
\end{align}
with $\alpha_t>0$ being the learning rate, $\hbQ^{t}$ being the estimate of the SA-VF at time-step $t$, and $[\hbQ^t]_{s,a}=[\hbQ^{t-1}]_{s,a}$ for all $(s,a)\neq (s_t,a_t)$. $Q$-learning enjoys convergence properties, as long as all $(s,a)$ pairs are visited infinitely often \cite{watkins1992q}. $Q$-learning updates one entry of the matrix $\hbQ^t \in \reals ^{C_\ccalS \times C_\ccalA}$ at a time. Since no constraints are imposed on the $Q$ matrix, the number of parameters to estimate is $C_\ccalS C_\ccalA$ and, as a result, $Q$-learning can take a long time (number of samples) to converge \cite{bertsekas1996neuro}.

\subsection{Matrix low-rank TD learning}

Our approach to making feasible nonparametric TD learning in scenarios where the values of $C_\ccalS$ and $C_\ccalA$ are large is to devise \emph{low-rank} schemes to learn the SA-VF. The motivation is threefold: i) substituting $\bbQ$ by a low-rank estimate $\hbQ$ reduces the degrees of freedom of the model, accelerating convergence and providing a trade-off between expressiveness and efficiency; ii) low-rank methods have been successfully adopted in other data-science problems and provide some degree of interpretability; and iii) as illustrated in the numerical section, we have empirically observed that in many RL environments, the optimal SA-VF can be approximated as low-rank and, more importantly, that the policies induced by the low-rank SA-VF estimates yield a performance very similar to the optimal ones.  While constraints of the form $\mathrm{rank}(\bbQ)\leq K$ can be added to the optimization in \eqref{E:Q_estimation_plain}, the main drawback is that the resultant optimization is computationally intractable since the rank constraint is NP-hard. 
In the context of low-rank optimization, a versatile approach 
is to implement a non-convex matrix factorization technique consisting of replacing $\bbQ$ by two matrices $\bbL \in \reals^{C_\ccalS \times K} $ and $\bbR  \in \reals^{K \times C_\ccalA}$, such that $\bbQ=\bbL \bbR$. This clearly implies that the rank of $\bbQ$ is upper-bounded by $K$, which must be selected based on either prior knowledge or computational constraints. The optimization problem becomes
\begin{eqnarray}\label{E:Q_factor_estimation_plain}
\hspace{-.35cm}&\{ \hbL,\hbR \} \!= \! \argmin_{\bbL,\bbR} \sum_{(s,a)\in \ccalM} (q_s^a - [\bbL\bbR]_{s,a})^2.\;\;&
\end{eqnarray}
Although this problem is non-convex, in the context of matrix completion, computationally efficient algorithms based on alternating-minimization algorithms have been shown to converge to local minima \cite{markovsky2012low,udell2016generalized}. Indeed, \eqref{E:Q_factor_estimation_plain} can be solved using Alternating Least Squares (ALS). However, this implies solving an iterative ALS problem for each sample pair $(s, a) \in \ccalM$, which is computationally expensive \cite{rozada2021low}.

In this work, we propose using a stochastic block-coordinate gradient descent algorithm to solve \eqref{E:Q_factor_estimation_plain}. Let $[\hbL^t]_{s_t}\in \reals^K$ be a vector collecting the $K$ entries of $\hbL^t$ associated with the state $s_t$. Similarly, let $[\hbR^t]_{a_t}\in \reals^K$ be a vector collecting the $K$ entries of $\hbR^t$ associated with the action $a_t$. 

\noindent \textbf{Proposed algorithm:} at time $t$, given the tuple $(s_t,a_t,{s_{t+1}})$, the stochastic update of $\hbL^t$ and $\hbR^t$ is done using the following schema:

\vspace{.15cm}
\noindent \begin{boxedminipage}[c]{88mm}
\noindent ~\emph{Step 1:} Update $~\forall~ [\bbs;\bba]\neq [\bbs^t;\bba^t]$
\begin{align}
[\hbL^t]_{s}~=&[\hbL^{t-1}]_{s}~\text{and}~[\hbR^t]_{a}=[\hbR^{t-1}]_{a}.\label{E:LR_learning_factor_alt_SGD_a}
\end{align}
\noindent ~\emph{Step 2a:} Run the stochastic update of the left matrix
\begin{align}
[\hbL^t]_{s_t} =&  [\hbL^{t-1}]_{s_t} + \alpha_t \big( r_t + \gamma \max_a [\hbL^{t-1}\hbR^{t-1}]_{s_{t+1},a}   \nonumber\\
-& [\hbL^{t-1}\hbR^{t-1}]_{s_t,a_t}\big)[\hbR^{t-1}]_{a_t}. \label{E:LR_learning_factor_alt_SGD_b}
\end{align}
\noindent ~\emph{Step 2b:} Run the stochastic update of the right matrix
\begin{align}
[\hbR^t]_{a_t} =&  [\hbR^{t-1}]_{a_t} + \alpha_t \big( r_t + \gamma \max_a [\hbL^{t}\hbR^{t-1}]_{s_{t+1},a}  \nonumber\\ -& [\hbL^{t}\hbR^{t-1}]_{s_t,a_t}\big)[\hbL^t]_{s_t}.
\label{E:LR_learning_factor_alt_SGD_c}
\end{align}
\noindent ~\emph{Step 3:} Generate the current estimate of the $Q$ matrix as
\begin{align}
\hbQ^t =&  \hbL^t\hbR^t.
\label{E:LR_learning_factor_alt_SGD_d}
\end{align}

\end{boxedminipage}
\vspace{.15cm}

This corresponds to a stochastic block-coordinate gradient descent algorithm where the value of the true SA-VF is estimated online as in \eqref{E:Q_learning_estimation}. Taking a closer look at the updates, we note that if a state is not observed at time $t$, \eqref{E:LR_learning_factor_alt_SGD_a} dictates that row ($K$ values) of $\hbL$ codifying that state remain unchanged. The same is true if an action is not taken at time $t$. Differently, \eqref{E:LR_learning_factor_alt_SGD_b} and \eqref{E:LR_learning_factor_alt_SGD_c} deal with the state and action associated with time instant $t$. In \eqref{E:LR_learning_factor_alt_SGD_b}, the $K$ values in $\hbL$ that codify the state $s_t$ are updated using an estimate of the gradient of \eqref{E:Q_factor_estimation_plain} with respect to (w.r.t.) $\hbL$. The structure of \eqref{E:LR_learning_factor_alt_SGD_c}, which updates the $K$ values in $\hbR$ that codify the action $a_t$, is similar to \eqref{E:LR_learning_factor_alt_SGD_b}, with the exception that, since the block of variables $\hbL$ was already updated, the $Q$-matrix is estimated as $\hbL^{t}\hbR^{t-1}$ in lieu of $\hbL^{t-1}\hbR^{t-1}$. 
While the spirit of \eqref{E:LR_learning_factor_alt_SGD_a}-\eqref{E:LR_learning_factor_alt_SGD_c} is similar to that of the $Q$-learning update in \eqref{E:Q_learning_estimation}, two main advantages of the matrix low-rank approach are: i) the number of parameters to estimate goes from $C_\ccalS C_\ccalA$ down to $(C_\ccalS + C_\ccalA)K$ and ii) at each $t$, the whole row $[\hbL^t]_{s_t}$ and column $[\hbR^t]_{a_t}$ (with $K$ values each) are updated. Meanwhile, in $Q$-learning only a single parameter $[\hbQ^t]_{s_t,a_t}$ is updated at time $t$. On the other hand, the best one can hope for this scheme is to converge to a local optimum, provided that $\alpha_t$ is sufficiently small, the rewards are bounded, and the observations are unbiased. Although the speed of convergence can be an issue, the stepsize and/or the update in \eqref{E:LR_learning_factor_alt_SGD_a}-\eqref{E:LR_learning_factor_alt_SGD_c} can be re-scaled using the norm of the stochastic gradient to avoid the potential stagnations caused by plateaus and saddle points commonly present in non-convex problems \cite{murray2019revisiting}.

\vspace{.1cm}
\noindent\textbf{Frobenious-norm regularization:} Alternatively, low-rankness in \eqref{E:Q_estimation_plain} can be promoted by using the nuclear norm $\|\bbQ\|_*$ as a \emph{convex} surrogate of the rank \cite{rozada2021low}. The nuclear norm is defined as $\|\bbQ\|_*=\sum_{k=1}^{K_{\max}} \sigma_k$, where $\sigma_k$ are the singular values of $\bbQ$ and $K_{\max}=\min\{C_\ccalS,C_\ccalA\}$ is the maximum rank. The nuclear norm $\|\bbQ\|_*$ is added to \eqref{E:Q_estimation_plain} as a regularizer, inducing sparsity on the singular values of $\bbQ$, and effectively reducing the rank \cite{markovsky2012low}. Solving nuclear-norm-based problems is challenging though, since interior-point methods are cubic on the number of elements in $\bbQ$ and more sophisticated iterative methods that implement a thresholding operation on the singular values require computing an SVD at every iteration \cite{burer2005local,markovsky2012low}. Luckily, the nuclear norm can also be written as $\|\bbQ\|_*=\frac{1}{2}\min_{\bbL\bbR=\bbQ} \|\bbL\|_F^2 + \|\bbR\|_F^2$ \cite{srebro2005rank}. This is important because we can then recast the nuclear-norm regularized estimation as
\begin{equation}\label{E:Q_factor_estimation_frobenius_norm}
      \{ \hbL,\hbR \} \!\!=\! \argmin_{\bbL,\bbR} \!\sum_{(s,a)\in \ccalM} \!(q_s^a - [\bbL\bbR]_{s,a})^2 + \eta \|\bbL\|_F^2 + \eta \|\bbR\|_F^2,
\end{equation}
which clearly resembles \eqref{E:Q_factor_estimation_plain}. As a result, we can handle \eqref{E:Q_factor_estimation_frobenius_norm} developing a stochastic algorithm similar to that in \eqref{E:LR_learning_factor_alt_SGD_a}-\eqref{E:LR_learning_factor_alt_SGD_c}. Algorithmically, adapting \eqref{E:LR_learning_factor_alt_SGD_a}-\eqref{E:LR_learning_factor_alt_SGD_c} to the cost in \eqref{E:Q_factor_estimation_frobenius_norm} requires accounting for the gradients of the Frobenius norms w.r.t. the rows of the factors. This implies adding $-\alpha_t\eta [\hbL^{t-1}]_s$ to the updates of the rows of $\hbL$ [cf. \eqref{E:LR_learning_factor_alt_SGD_a} and \eqref{E:LR_learning_factor_alt_SGD_b}] and adding $-\alpha_t\eta [\hbR^{t-1}]_a$ to the updates of the rows of $\hbR$ [cf. \eqref{E:LR_learning_factor_alt_SGD_a} and \eqref{E:LR_learning_factor_alt_SGD_c}]. Intuitively, when \eqref{E:Q_factor_estimation_plain} is replaced with \eqref{E:Q_factor_estimation_frobenius_norm}, one can increase the value of $K$ (size of matrices $\bbL$ and $\bbR$). Overshooting for the value of $K$ is less risky in this case since the role of the Frobenious regularizers is precisely keeping the effective rank under control. It is also important to remark that, since the nuclear norm is convex, the stochastic schemes optimizing \eqref{E:Q_factor_estimation_frobenius_norm} are expected to have its behavior stabilized as well. The main drawback is that, since larger $K$'s are typically allowed, the number of samples that the stochastic schemes require to converge increases as well. Interestingly, our simulations show that, when considering practical considerations such as training time and sample complexity, the reward achieved by the Frobenious-regularized schemes is not significantly different from the non-regularized ones. Hence, for the sake of readability and concision, we will focus most of our discussion on the simpler schemes in \eqref{E:LR_learning_factor_alt_SGD_a}-\eqref{E:LR_learning_factor_alt_SGD_c}.

\subsection{Low-rank SA-VF approximation meets linear methods}

As pointed out previously, linear parametric schemes can be alternatively used to estimate the SA-VF. To facilitate the discussion, let us suppose that the number of features considered by the model is $K$ and remember that linear schemes model the SA-VF as $[\hbQ]_{s,a}\!=\!Q_{\beta}(s, a)\!=\!\bbphi(s, a)^\top \bbbeta$, with $\bbphi(s, a)\in \reals^K$ being the features that characterize the state-action pair and $\bbbeta\in \reals^K$ are the parameters to be estimated. A central problem of such methods is to design the feature vectors $\bbphi(s, a)$. This is carried out using prior knowledge or imposing desirable properties on $\bbphi(s,a)$ and, oftentimes, is more an art than a science. Differently, the approach proposed in the previous sections amounts to learning the feature vectors and the parameters \emph{jointly}. More precisely, if $\text{rank}(\hbQ)\!=\!K$, the SVD can be used to write $\hbQ\!=\!\bbV \diag(\bbsigma) \bbU^\top$, where $\bbV\!=\![\bbv_1,...,\bbv_K]$ and $\bbU\!=\![\bbu_1,...,\bbu_K]$ are the left and right singular vectors, and $\bbsigma\!=\![\sigma_1,...,\sigma_K]^\top$ the singular values. The SVD can be reformulated as $\hbQ=\sum_{k=1}^K \sigma_k \bbv_k \bbu_k^\top$, which in turn implies that $[\hbQ]_{s,a}=\sum_{k=1}^K \sigma_m [\bbv_k]_s [\bbu_k]_a$. The low-rank approach implemented here can be related to a linear model where the features are  $\bbphi(s, a) = [[\bbv_1]_s [\bbu_1]_a,..., [\bbv_K]_s [\bbu_K]_a]^\top$ and the parameters as $\bbbeta = \bbsigma$, with the difference being that here both $\bbphi(s, a)$ and $\bbbeta$ are learned together in an online fashion by interacting with the environment. As a reminder, there are alternative methods that also address the challenge of requiring manual feature engineering, NNs being a prominent example \cite{goodfellow2016deep}.

\section{Tensor low-rank for SA-VF estimation}\label{S:low_rank_tensor_Q_estimation}

In most practical scenarios, both the state and the action spaces are composed of various dimensions. In this section, we put forth a novel modeling approach that \textit{represents the SA-VF as a tensor}, which accounts naturally for those multiple dimensions. To be rigorous, let us suppose that the state space $\ccalS$ has $D_\ccalS$ state dimensions and the action space $\ccalA$ has $D_\ccalA$ action dimensions. We can then use the vector $\bbs=[s_1, ..., s_{D_\ccalS}]^\top$ to represent the state of the system and $\bba=[a_1, ..., a_{D_\ccalA}]^\top$ to represent the action to take. Let $\ccalS_d$ denote the set of values that $s_d$ can take and $C_{\ccalS_d}$ the cardinality of that set. Similarly, we use $\ccalA_d$ and $C_{\ccalA_d}$ for all $d=1,...,D_\ccalA$ to denote their counterparts in the action space. Using the Cartesian product, the state space and the action space are defined as 
\begin{equation}\label{E:space_action_spaces_as_Cartesian_products}
\ccalS=\ccalS_1 \times ... \times\ccalS_{D_\ccalS}~\text{and}~\ccalA=\ccalA_1 \times ... \times\ccalA_{D_\ccalA},
\end{equation}
respectively. Moreover, their respective cardinalities satisfy 
\begin{equation}
C_\ccalS=\prod_{d=1}^{D_\ccalS}C_{\ccalS_d}~\text{and}~C_\ccalA=\prod_{d=1}^{D_\ccalA}C_{\ccalA_d}.
\end{equation}
Clearly, as the number of dimensions increases, the number of values of the SA-VF to estimate grows exponentially. More importantly for the discussion at hand, classical RL estimation methods arrange the SA-VF in the form of a matrix $\bbQ$. In these cases, the rows of the $Q$-matrix index the Cartesian product of all one-dimensional state sets, and columns do the same for the Cartesian product of the actions. As a result, even if we use a low-rank matrix approximation, the number of parameters to estimate grows exponentially with $D_\ccalS$ and $D_\ccalA$


In scenarios where the states and actions have multiple dimensions, tensors are a more natural tool to model the SA-VF. This multi-dimensional representation of the SA-VF will be referred to as $Q$-tensor and denoted as $\tenbQ$. This tensor has $D=D_\ccalS+D_\ccalA$ dimensions, each of them corresponding to one of the individual (space or action) sets in \eqref{E:space_action_spaces_as_Cartesian_products}. To simplify notation, in this section we combine the state and action states and, to that end, we define joint state-action space $\ccalD=\ccalS \times \ccalA$, that has $D$ dimensions; introduce the $D$-dimensional vector $[\bbs;\bba]=[\bbs^\top,\bba^\top]^\top$ to denote an element of $\ccalD$; and use $C_d$ to denote the number of state (action) values along the $d$th dimension of the tensor, so that $C_d=|\ccalS_d|$ if $1\leq d\leq D_\ccalS$ and $C_d=|\ccalA_{d-D_\ccalS}|$ if $D_\ccalS<d\leq D_\ccalS + D_\ccalA$. With this notation at hand, it readily follows that $\tenbQ$ is a tensor (multi-array) with $D$ dimensions, with size $C_1\times ... \times C_D$, where each of its $\prod_{d=1}^D C_d$ entries can be indexed as $[\tenbQ]_{[\bbs;\bba]}$.


Similarly to other multi-dimensional applications \cite{kolda2009tensor}, \cite{sidiropoulos2017tensor}, low-rank tensor factorization techniques can be adapted to impose low-rank in the estimation of $\tenbQ$, which is the goal of the remainder of the section. To that end, Section \ref{Ss:tensor_parafac_generic} introduces the notion of PARAFAC low-rank decomposition, Section \ref{Ss:Belman_tensor_lowrank} briefly describes the synchronous model-based low-rank tensor SA-VF setup, Section \ref{Ss:tensor_lowrank_stochastic_schemes} formulates the low-rank tensor-based SA-VF for the main model-free setup and presents the baseline version of our stochastic algorithm, and Sections \ref{Ss:tensor_stachastic_title_to_decide} and \ref{Ss:matrix_and_tensor_stochastic} and discuss potential modifications to the algorithms at hand. 

\subsection{PARAFAC decomposition}\label{Ss:tensor_parafac_generic}
While multiple definitions for the rank of a tensor exist, our focus is on the one that defines the rank of a tensor as the minimum number of \emph{rank-1} factors that need to be combined (added) to recover the whole tensor. Adopting such a definition implies that a tensor $\tenbX$ of rank $K$ can be decomposed as $\tenbX = \sum_{k=1}^K \tenbX^k$. \emph{Rank-1} matrices can be expressed as the outer product of two vectors $\bbX=\bbl \bbr^\Tr=\bbl \circledcirc \bbr$. Similarly, a \emph{rank-1} tensor of order $D$ can be expressed as the outer product of $D$ vectors $\tenbX_{rank-1}=\bbx_1\circledcirc...\circledcirc\bbx_D$. The decomposition of a tensor as the sum of the outer product of vectors 
$$\tenbX = \sum_{k=1}^K \bbx_1^k\circledcirc...\circledcirc\bbx_D^k$$ 
is known as the PARAFAC decomposition \cite{bro1997parafac}. Upon defining the  matrices $\bbX_d=[\bbx_d^1,...,\bbx_d^K]$, known as factors, for all $d=1,...,D$, the PARAFAC decomposition can be matricized along each dimension, or mode, as 
$$\mathrm{mat}_d(\tenbX) = (\bbX_1 \odot...\odot \bbX_{d-1} \odot \bbX_{d+1} \odot ... \odot \bbX_D) \bbX_d^\Tr,$$ 
where $\odot$ denotes the \emph{Khatri-Rao} or \emph{column-wise Kronecker} product. Mathematically, we define the matricization operator $\mathrm{mat}_d(\cdot)$ as

\begin{align}
    \bbX=&\mathrm{mat}_{d}(\tenbX)\in\reals^{(C_1...C_{d-1}C_{d+1}...C_D)\times C_d}\;\mathrm{where}\hfill\\
     &[\bbX]_{k,n_d}=[\tenbX]_{n_1,...,n_D} \;\text{and}\; \nonumber\\
     &k = n_1+\sum_{i=2,i\neq d}^{I}(n_i-1)\prod_{j=2,j\neq d}^{i-1}C_j.\nonumber
\end{align}
where $d\leq D$ and, to simplify exposition, we have assumed that $d>1$.

\noindent To simplify ensuing mathematical expressions, we define: a) $\bigodot_{i \neq d}^D \bbX_i:=\bbX_1 \odot...\odot \bbX_{d-1} \odot \bbX_{d+1} \odot ... \odot \bbX_D$ and b) $\mathrm{unmat}_d(\cdot)$ as the inverse operator of $\mathrm{mat}_d(\cdot)$, that is, the one that for the matrix input $\mathrm{mat}_d(\tenbX)$ generates the tensor $\tenbX$ as the output.

Finally, one can be interested in approximating a tensor $\tenbX$  by a tensor of rank $K'$. With $[i_1,...,i_D]$ 
being an index vector and recalling that $\tenbX_{[i_1,...,i_D]}$ denotes a particular entry of the tensor $\tenbX$, we define $\hat{\tenbX}$, the rank $K'$ approximation of $\tenbX$, as
\begin{eqnarray}
    \nonumber   \hat{\tenbX}&=&\mathrm{PARAFAC}_{K'}(\tenbX):=\argmin_{\smalltenbZ~} \sum_{\forall [i_1,...,i_D]} \Big([\tenbX]_{[i_1,...,i_D]} \\
   &-& [\tenbZ]_{[i_1,...,i_D]}\Big)^2~\mathrm{subject~to}:~\mathrm{rank}(\tenbZ)\leq K',\label{E:parafac_approx}
\end{eqnarray}
which serves as the counterpart to the TSVD operator in \eqref{E:truncatedSVD_closedformRank1} for the tensor case.  

\subsection{Tensor low-rank model in Bellman's equations}\label{Ss:Belman_tensor_lowrank}

As done for the matrix case, the first step is to impose low rank of the  tensor representation of the SA-VF. To that end, we leverage the truncated PARAFAC approximation in \eqref{E:parafac_approx} and propose the following iteration 
\begin{equation}\label{E:Bellman_parafac_low_rank}
 \hbq^\pi_{t+1} = \mathrm{vec} (\mathrm{PARAFAC}_K (\mathrm{unvec}(\bbr +\gamma \bbP \bbPi \hbq^\pi_{t})),
 \end{equation} 
which is the counterpart to \eqref{E:Bellman_low_rank} for the tensor case. Note that with a slight abuse of notation, in the equation above i) $\mathrm{vec}$ denotes the vectorization operator that maps a (high-dimensional) tensor into a column vector stacking all its fibers ii)  and $\mathrm{unvec}$ denotes the inverse operator that maps the stacked vector back to the dimensions of the original tensor. Interestingly, the scheme \eqref{E:Bellman_parafac_low_rank} also converges to a neighborhood of $\bbq^\pi$, as formally stated next.  To see this, we need first to introduce the following straightforward lemma.






     

%
\begin{lemma}
 \label{lemma_parafac}
 Let $\tenbQ \in \reals^{N_1 \times \ldots \times N_D}$ be a tensor whose PARAFAC decomposition is given by 
 $\tenbQ =\sum_{i=1}^{R} \tenbQ^i$ with $\tenbQ^i=\bbq_1^i \circledcirc \ldots \circledcirc \bbq_D^i$. Assume, without loss of generality, that the rank-1 factors are ordered so that $\|\mathrm{vec}(\tenbQ^i)\|_\infty\geq\|\mathrm{vec}(\tenbQ^{i+1})\|_\infty$. Then, upon defining $\xi_i:=\|\mathrm{vec}(\tenbQ^i)\|_\infty$, it holds that 
  \begin{equation}
     \left\| \mathrm{vec}(\mathrm{PARAFAC}_{K'}(\tenbQ)) - \mathrm{vec}(\tenbQ) \right\|_\infty \leq (R - K') \xi_{K'+1}.
 \end{equation}
 \end{lemma}

\begin{proof}
Since both terms inside the norm share the first $K'$ rank-1 components, leveraging the triangular inequality yields

     \begin{align}
         &\| \mathrm{vec}(\mathrm{PARAFAC}_{K'}(\tenbQ)) - \mathrm{vec}(\tenbQ) \|_\infty = \left\|\sum_{i=K' + 1}^{R}  \mathrm{vec}(\tenbQ^i)     \right\|_\infty \nonumber\\
         &\leq \sum_{i=K' + 1}^{R} \left\| \mathrm{vec}(\tenbQ^i) \right\|_\infty=\sum_{i=K' + 1}^{R} \xi_i. \label{eq:BoundLemma1}
     \end{align}

     \noindent Then, leveraging that $\max \{\xi_i\}_{i=K'+1}^R=\xi_{K'+1}$, we conclude that the left hand side of \eqref{eq:BoundLemma1} is bounded by $(R - K')\xi_{K'+1}$, finishing the proof.


     
 \end{proof}

Then, we have the following corollary:

\begin{corollary}\label{corollary}
Let $K'>0$ be a constant denoting the rank of the approximation operator $\mathrm{PARAFAC}_{K'}(\cdot)$. Moreover, given the $D$-dimensional tensor $\mathrm{unvec}(\bbr + \gamma \bbP \bbPi \hbq^\pi_{t})$, let $N_d$ denote the length its $d$th dimension and $\xi_{K'+1}(t)$ the infinite norm of its $(K'+1)$th rank-1 factor. Then, with $B'$ being a constant defined as

\begin{eqnarray}
    \label{eq::b_prime_def}
    \hspace{-0.4cm}&B' \!= \!\max_{t=1,2,...}\!\left(\!\min\!\left\{ \prod_{d = 1 \neq i}^D N_d\right\}_{i=1}^D\!\!\!\!-K'\!\!\right)\! \xi_{K'+1}(t),&
\end{eqnarray}

\noindent it holds that the estimate in \eqref{E:Bellman_parafac_low_rank}  satisfies 
\begin{equation}
\limsup_{t\to\infty}\left\|\hat{\bbq}_t^\pi - \bbq^\pi \right\|_\infty\leq \frac{B'}{1-\gamma}.
\end{equation}
\end{corollary}

Both, the corollary and its proof are almost identical to Proposition \ref{theorem}. In fact, the only difference between them is the characterization of the constant $B'$. In this case, the infinity-norm error between the application of the Bellman operator and the truncated PARAFAC depends on the truncated rank-1 components of the tensor $\mathrm{unvec}(\bbr + \gamma \bbP \bbPi \hbq^\pi_{t})$, which, as in the SVD case, is always bounded.

\subsection{Tensor low-rank TD learning}\label{Ss:tensor_lowrank_stochastic_schemes}
The next step is to move to more realistic model-free RL scenarios. This requires adapting the formulation in \eqref{E:Q_estimation_plain} to the tensor low-rank scenario considered in this section. Leveraging the tensor notation and concepts just introduced, this yields

\begin{eqnarray} \label{E:Q_tensor_estimation_plain}
\hat{\tenbQ} &=& \argmin_{\smalltenbQ} \sum_{(\bbs, \bba)\in \ccalM} (q_\bbs^\bba - [\tenbQ]_{[\bbs; \bba]})^2\\
&~&\text{subject~to}:~\mathrm{rank}(\tenbQ)\leq K. \label{E:Q_tensor_estimation_plain_constraint}
\end{eqnarray}
Since we adopt the  PARAFAC decomposition, the constraint in \eqref{E:Q_tensor_estimation_plain_constraint} can be alternatively written as
\begin{equation}\label{E:Q_tensor_estimation_plain_constraint_v2}
\hat{\tenbQ} = \sum_{k=1}^K \hbq_1^k\circledcirc...\circledcirc\hbq_D^k    
\end{equation}
with the factor matrices $\hbQ_d=[\hbq_d^1,...,\hbq_d^K]\in \reals^{C_d\times K}$ being the effective optimization variables. The products in \eqref{E:Q_tensor_estimation_plain_constraint_v2} render the optimization non-convex. As in \eqref{E:Q_factor_estimation_plain}, the approach to developing an efficient algorithm is to optimize over each of the $D$ factors iteratively. To be more specific, let us suppose that the factor matrices are known (or fixed) for all $i\neq d$, then the optimal factor matrix for the $d$th dimension can be found as  
\begin{equation}
    \label{E:parafac}
    \hbQ_d=\argmin_{\bbQ_d}\!\!\sum_{(\bbs, \bba)\in \ccalM}\! \Big(q_\bbs^\bba - \Big[\mathrm{unmat}_d\Big(\Big(\bigodot_{i \neq d}^D \hbQ_i\Big) \bbQ_d^\Tr \Big)\Big]_{[\bbs; \bba]}\Big)^2,
\end{equation}
which is a classical least squares problem. While the original PARAFAC tensor decomposition is amenable to alternating-minimization schemes, we remark that: i) the cost in \eqref{E:Q_tensor_estimation_plain} only considers a collection of sample state-action pairs $\ccalM$ from the environment; and ii) our main interest is in online RL environments where the samples are acquired on the fly. As a result, stochastic block-coordinate gradient descent algorithms are a more suitable alternative to solve \eqref{E:Q_tensor_estimation_plain}-\eqref{E:Q_tensor_estimation_plain_constraint}.

To describe our proposed algorithm in detail, we need first to introduce the following notation:
\begin{itemize}

\item $\hat{\tenbQ}^t$ is the estimate for the state-action value tensor of $D=D_\ccalS+D_\ccalA$ dimensions generated for time instant $t$. 

\item $\hbQ_d^t \in \reals^{|\ccalD_d|\times K}$ is the factor matrix of the $d$th dimension of $\ccalD$ at time-step $t$. 

\item $\bbs^t=[s_1^t, ..., s_{D_\ccalS}^t]^\top$ is the instantaneous state vector, $\bba^t=[a_1^t, ..., a_{D_\ccalA}^t]^\top$ the instantaneous action vector, and $[\bbs^t;\bba^t]$ the $D$-dimensional state-action vector.

\item $[\hbQ_d^t]_{[\bbs^t;\bba^t]_d} \in \reals^K$ is a column \emph{vector} whose entries represent the row of matrix $\hbQ_d^t$ indexed by the $d$th entry of the instantaneous state-action vector $[\bbs^t;\bba^t]$. 

\item $\big[\hbQ_{\backslash d}^t\big]_{[\bbs^t;\bba^t]} \in \reals^K$ is a column \emph{vector} whose $k$th entry is defined as 
        $$\Big[\big[\hbQ_{\setminus d}^t\big]_{[\bbs^t;\bba^t]}\Big]_k = \!\prod_{i=1}^{d\!-\!1} \!\Big[[\hbQ_{i}^t]_{[\bbs^t;\bba^t]_i}\Big]_k\!\prod_{i=d\!+\!1}^{D}\!\Big[[\hbQ_{i}^{t-1}]_{[\bbs^t;\bba^t]_i}\Big]_k,$$
that is, we generate the vector $[\hbQ_{\backslash d}^t\big]_{[\bbs^t;\bba^t]}$ as the Hadamard product of $D-1$ vectors. 
\item  $\hat{\tenbQ}_d^{t-1}$ is a tensor with the same dimensions as $\hat{\tenbQ}^t$ computed from the factors $\{\hbQ_i^t\}_{i=1}^d$ and $\{\hbQ_i^t\}_{i=d+1}^D$ as 
        $$\hat{\tenbQ}_d^{t-1} = \mathrm{unmat}_d\left(\left(\bigodot_{i=1}^{d-1} \hbQ_i^{t} \bigodot_{i = d+1}^D \hbQ_i^{t-1}\right) (\hbQ_d^{t-1})^\top\right)$$ 

\end{itemize}

\vspace{.1cm}
\noindent\textbf{Proposed algorithm:} Leveraging these notational conventions, at time $t$ and given the values of $\bbs^t$, $\bba^t$ and $\bbs^{t+1}$, the proposed stochastic update to estimate $\hat{\tenbQ}^t$ proceeds as follows. 

\vspace{.15cm}
\noindent \begin{boxedminipage}[l]{88mm}
\noindent ~\emph{Step 1:} For $d=1,...,D$ update
\begin{align}
[\hbQ_d^t]_{[\bbs;\bba]_d} =& [\hbQ_d^{t-1}]_{[\bbs;\bba]_d}~\forall~ [\bbs;\bba]_d\neq [\bbs^t;\bba^t]_d.\label{E:tensor_learning_factor_alt_SGD_a}
\end{align}
\noindent ~\emph{Step 2:} For $d=1,...,D$ run the stochastic updates
\begin{align}\label{E:tensor_learning_factor_alt_SGD}
[\hbQ_d^t]_{[\bbs^t;\bba^t]_d} =&  [\hbQ_d^{t-1}]_{[\bbs^t;\bba^t]_d} + \alpha_t \big( r_t + \gamma \max_\bba [\hat{\tenbQ}_d^{t-1}]_{[\bbs^{t+1};\bba]} \nonumber \\
-& [\hat{\tenbQ}_d^{t-1}]_{[\bbs^t;\bba^t]}\big) \big[\hbQ_{\backslash d}^t\big]_{[\bbs^t;\bba^t]}. 
\end{align}
\noindent ~\emph{Step 3:} Set
\begin{align}
\hat{\tenbQ}^{t} =&  \mathrm{unmat}_D\left(\left(\bigodot_{i=1}^{D-1} \hbQ_i^{t} \right) (\hbQ_D^{t})^\top\right).\label{E:tensor_learning_factor_alt_SGD_c}
\end{align}
\end{boxedminipage}
\vspace{.25cm}

\noindent The structure of this stochastic update is related to that of the matrix low-rank TD update in \eqref{E:LR_learning_factor_alt_SGD_a}-\eqref{E:LR_learning_factor_alt_SGD_c}, but here we consider $D$ different blocks of variables while in the matrix case there were only two. Each block of variables corresponds to one of the factor matrices $\hbQ_d$ (all with $K$ columns), which are then updated sequentially (for $d=1,...D$) running a stochastic gradient scheme. More specifically, Step 1 considers the rows of the factors  associated with the states and actions not observed at time $t$. Since no new information has been collected, the estimates for $t$ are the same as those for the $t-1$. Then, for each factor $\hbQ_d$ with $d=1,..,D$ the ($K$ values of the) row associated with the observed state or action is updated in Step 2. As an illustration, if at time $t$ the index of the state $s_1^t$ is 3, then the third row of the matrix factor $\hbQ_1^t$ is updated. The update in \eqref{E:tensor_learning_factor_alt_SGD} is similar to that in the classical TD $Q$-learning scheme with $\big[\hbQ_{\backslash d}^t\big]_{[\bbs^t;\bba^t]} $ in \eqref{E:tensor_learning_factor_alt_SGD} being a scaled version of the \emph{gradient} of the quadratic cost in \eqref{E:Q_tensor_estimation_plain} \emph{w.r.t. the row} $[\bbs_t,\bba_t]_d$ \emph{of the} $d$th \emph{matrix factor} $\hbQ_d$. Finally, the last step just combines all the matrix factors to generate the new tensor estimate for time $t$.


\vspace{.1cm}
\noindent\textbf{Scalability:} The schemes in \eqref{E:tensor_learning_factor_alt_SGD_a}-\eqref{E:tensor_learning_factor_alt_SGD_c} convey a reduction in the number of parameters needed to approximate the SA-VF. While the number of parameters of the matrix low-rank model is $(C_\ccalS + C_\ccalA)K =(\prod_{i=1}^{D_\ccalS} C_i + \prod_{i=D_\ccalS+1}^{D_\ccalS+D_\ccalA} C_i)K$, the tensor low-rank model becomes additive in the dimensions of the state-action space, leading to a total number of parameters of 
\begin{equation}
\left(\sum_{i=1}^{D_\ccalS} C_i + \sum_{i=D_\ccalS+1}^{D_\ccalS+D_\ccalA} C_i\right)K=\left(\sum_{i=1}^{D} C_i \right)K.
\end{equation}
If we assume for simplicity that $D_\ccalS=D_\ccalA=D/2$ and $C_d=C$ for all $d$, the ratio between the number of parameters for the tensor and matrix schemes is 
\begin{eqnarray}\label{E:numberofparams_matrix_vs_tensor}
\frac{(\sum_{i=1}^{D} C_i)K'}{(\prod_{i=1}^{D_\ccalS} C_i + \prod_{i=D_\ccalS+1}^{D_\ccalS+D_\ccalA} C_i)K} &=&\frac{D C K'}{(C^{D/2} + C^{D/2})K}\nonumber \\
&=& \frac{D K'}{2 K} C^{1-D/2},
\end{eqnarray}
demonstrating the significant savings of the tensor-based approach. Note that in the previous expression, we have considered that $K'$, the rank of the tensor, and $K$, the rank of the matrix, need not to be the same, since one expects higher rank values being used for the tensor case. 

The low-rank tensor structure is also beneficial from the point of view of accelerating convergence. As already discussed, one of the main drawbacks of the classical $Q$-learning algorithm is that, at each time instant $t$, only one entry of the SA-VF $Q$-matrix is updated. For the matrix low-rank TD algorithm, $C_\ccalS+C_\ccalA-1\approx C_\ccalS+C_\ccalA $ entries of the $Q$-matrix (those corresponding to the row indexing the current state and the column indexing the current action) were updated. To better understand the situation in the tensor case, let us focus on a particular dimension $d$. Then, it turns out that updating one of the rows of $\hbQ_d$ (the one associated with the index of the active state/action along the $d$th dimension) affects the value of $\prod_{i=1,i\neq d}^{D} C_i$ entries of the $Q$-tensor [cf. the Khatri-Rao matrix in \eqref{E:tensor_learning_factor_alt_SGD_c}]. Summing across the $D$ dimensions and ignoring the overlaps, the number of entries updated at each time instant is approximately $\sum_{d=1}^D \prod_{i=1, i\neq d}^D C_i$. 
If we assume once again for simplicity that $D_\ccalS=D_\ccalA=D/2$ and $C_d=C$ for all $d$, the ratio between the number of updated entries for the tensor and the matrix is 
\begin{equation}
\frac{\sum_{d=1}^D \prod_{i=1, i\neq d}^D C}{\prod_{i=1}^{D_\ccalS} C +  \prod_{i=1}^{D_\ccalA} C}=\frac{D C^{D-1}}{C^{D/2}+C^{D/2}}=\frac{D}{2}C^{D/2-1},
\end{equation}
demonstrating a substantial difference between the two approaches.

\subsection{Modifying the tensor low-rank stochastic algorithms} \label{Ss:tensor_stachastic_title_to_decide}

As discussed in Section \ref{S:low_rank_matrix_Q_estimation}, the TD algorithm in \eqref{E:tensor_learning_factor_alt_SGD_a}-\eqref{E:tensor_learning_factor_alt_SGD_c} can be viewed as a baseline that can be modified based on the prior information and particularities of the setup at hand. For example, if an adequate value for the rank is unknown and computational and sample complexity are not a limitation, one can select a high value for $K$ and regularize the problem with a term $\eta \|\hbQ_d\|_F^2$ that penalizes the Frobenious norm of each of the factors $d=1,...,D$. As in the case of the nuclear norm, this will promote low rankness so that the effective rank of the tensor will be less than $K$. Algorithmically, augmenting the cost in \eqref{E:Q_tensor_estimation_plain} with $\eta \sum_{d=1}^D \|\hbQ_d\|_F^2$ entails: i) considering the gradient of the regularizer w.r.t. rows of the factors $\hbQ_d$ and ii) adding such a gradient to the updates in Steps 1 and 2. The final outcome is the  addition of the term $-\eta [\hbQ_d^{t-1}]_{[\bbs;\bba]_d}$ to \eqref{E:tensor_learning_factor_alt_SGD_a} and the addition of the term $-\eta [\hbQ_d^{t-1}]_{[\bbs^t;\bba^t]_d}$ to \eqref{E:tensor_learning_factor_alt_SGD}.

In a different context, if the algorithm in \eqref{E:tensor_learning_factor_alt_SGD_a}-\eqref{E:tensor_learning_factor_alt_SGD_c} suffers from numerical instabilities associated with the non-convex landscape of the optimization problem, a prudent approach is to re-scale the stepsize with the norm of the stochastic gradient, which contributes to avoid a slow and unstable convergence \cite{murray2019revisiting}.

\subsection{Trading off matrix and tensor low-rank representations} \label{Ss:matrix_and_tensor_stochastic}

As explained in the previous sections, not all low-rank approaches are created equal. For the matrix case, the rows of $\bbL$ index every possible element of $\ccalS$ and the columns of $\bbR$ every possible element of $\ccalA$. Consideration of the simplest low-rank model in that case,  i.e., setting the rank to 1, requires estimating $2C^{D/2}$ parameters [cf. \eqref{E:numberofparams_matrix_vs_tensor}], which may not be realistic if $D$ is too large. Differently, the rank-1 model for the tensor case requires estimating only $DC$ parameters, which is likely to lead to poor estimates due to a lack of expressiveness. One way to make those two models comparable in terms of degrees of freedom would be to consider a very high rank (in the order of $C^{D/2-1}$) for the tensor approach, but this could result in numerical instability potentially hindering the convergence of the stochastic schemes. 


A different alternative to increase the expressiveness while keeping the rank under control is to group some of the $D$ dimensions of $\tenbQ$ and define a lower-order tensor representation. To be more specific, let us assume that $D_\ccalS$ and $D_\ccalA$ are even numbers and consider the set 
\begin{eqnarray}
\ccalP&=&\big\{ \{ \ccalS_1 \times... \times  \ccalS_{D_\ccalS/2}\}, \{ \ccalS_{D_\ccalS/2+1} \times... \times  \ccalS_{D_\ccalS}\},\nonumber\\
&~& \{ \ccalA_1 \times... \times  \ccalA_{D_\ccalA/2}\},  \{ \ccalA_{D_\ccalA/2+1} \times... \times  \ccalA_{D_\ccalA}\}\big\}.\nonumber
\end{eqnarray}
We can now define a tensor $\tenbQ$ with $4$ dimensions, each associated with one of the elements of $\ccalP$. Alternatively, we could consider the set 
\begin{eqnarray}
\ccalP'&=&\big\{ \{ \ccalS_1 \times \ccalS_2\},\{ \ccalS_3 \times \ccalS_4\},..., \{\ccalS_{D_\ccalS-1 }\times  \ccalS_{D_\ccalS}\},\nonumber\\
&~& \{ \ccalA_1 \times  \ccalA_2\},\{ \ccalA_3 \times  \ccalA_4\},..., \{ \ccalA_{D_\ccalA-1} \times \ccalA_{D_\ccalA}\}\big\},\nonumber
\end{eqnarray}
which would lead to a tensor $\tenbQ$ with $D/2$ dimensions. A rank-1 representation for the tensor based on $\ccalP$ would require estimating $4C^{D/4}$ parameters, while for the tensor based on $\ccalP'$ the rank-1 representation would require estimating $\frac{D}{2}C^{2}$ values. By playing with the number of dimensions and how the individual sets are grouped (including going beyond regular groups), the benefits of the matrix and tensor-based approaches can be nicely traded off. Note that one particular advantage of this clustering approach is that the different dimensions can be grouped based on prior knowledge. For example, the groups can be formed based on how similar the dimensions are. Alternatively, one can rely on operational constraints and group dimensions based on their domain (e.g., by grouping binary variables together, separating them from real-valued variables). The next section will explore some of these alternatives, showcasing the benefits of this approach. 


\section{Numerical experiments}

We test the algorithms in various standard RL problems, using the toolkits OpenAI Gym  \cite{brockman2016openai} \cite{Goddard2021} and Gym Classics \cite{daley2021gym}, and the Highway environments in \cite{highway-env}. As detailed in each of the experiments, the reward functions of some of the environments have been modified to penalize ``large'' actions, promoting agents to behave more smoothly. Furthermore, some of them have also been modified to have a continuous action space. The full implementation details of the environments can be found in \cite{Rozada2023}. Also, the Tensorly Python package has been used to handle multi-dimensional arrays and perform tensor operations \cite{kossaifi2016tensorly}. To illustrate and analyze the advantages of the matrix low-rank (MLR) and tensor low-rank (TLR) algorithms we i) study the effects of imposing low-rankness in the SA-VF estimated by classic algorithms, ii) assess the performance of the proposed algorithms in terms of cumulative reward, iii) compare the efficiency of the proposed algorithms in terms of number of parameters and speed of convergence against standard baselines, and iv) test TLR in a particularly challenging high-dimensional environment.

\subsection{Low rank in estimated SA-VF}

One of the arguments used when motivating our schemes was the existence of practical environments where the SA-VF matrix is approximately low rank. We start our numerical analysis by empirically testing this claim in different scenarios.

\begin{figure}
    \centering
    \includegraphics[width=87mm]{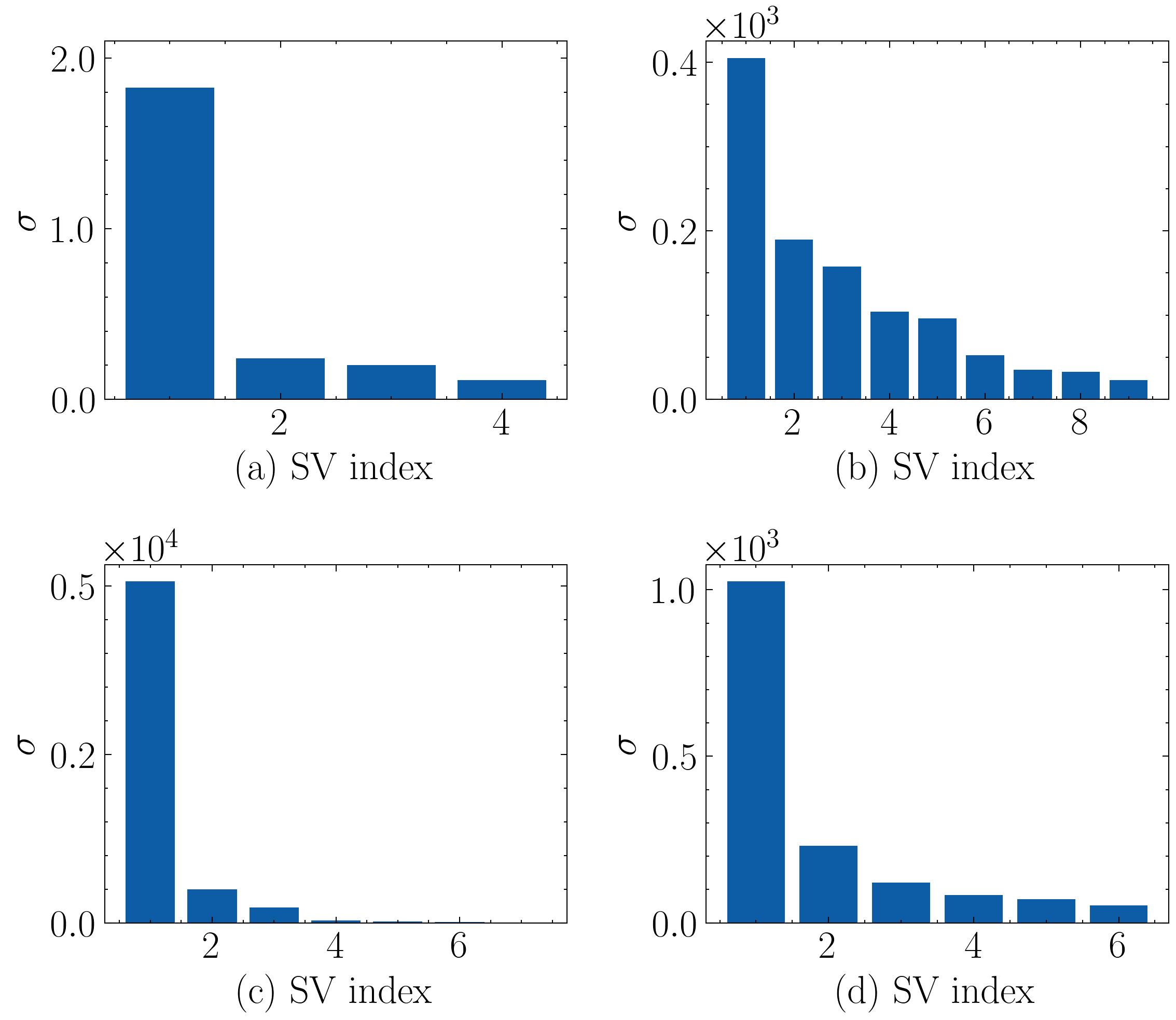}
    \caption{Singular values (SV) of $\bbQ^*$ obtained via policy iteration in four standard RL problems: (a) Frozenlake, (b) Racetrack, (c) JacksCarRental, and (d) Taxi-v3. The energy of $\bbQ^*$ tends to concentrate on the main SVs.}
    \label{fig:example_true}
\end{figure}

First, we analyze the optimal SA-VF matrix $\bbQ^*$ of some classic RL environments. Assuming that we have access to the reward vector $\bbr$, the transition probability matrix $\bbP$, and the policy matrix $\bbPi$, we can use the Bellman equation defined in \eqref{E:Belman_Qvectorized} to obtain the true $\bbq^\pi$ of a fixed policy $\pi$. As stated previously, this problem is known as \textit{policy evaluation}. Backed by the policy improvement theorem \cite{bertsekas2019reinforcement}, a new policy $\pi'$ can be obtained from $\bbq^\pi$ by solving $\pi'(s)=\argmax_a Q(s, a)$ for every single state $s$. This problem is known as \textit{policy improvement}. The policy iteration algorithm \cite{bertsekas2019reinforcement} repeats these two steps iteratively until the optimal SA-VF matrix $\bbQ^*$ is reached. Once the optimal $\bbQ^*$ is obtained, we can compute its SVD decomposition and study the magnitude of its singular values, assessing whether $\bbQ^*$ matrix is approximately low-rank or not. The singular values $\bbsigma$ of the matrix $\bbQ^*$ obtained for various simple MDP scenarios are reported in  Fig. \ref{fig:example_true}. The results reveal that just a few singular values of the matrices $\bbQ^*$ are large, while most of them are small. This illustrates that the ground-truth SA-VF resulting from these problems can be well approximated (in a Frobenius sense) by a low-rank matrix, that is, by a sum of just a few parsimonious rank-1 matrices.

\begin{figure}
    \centering
    \includegraphics[width=87mm]{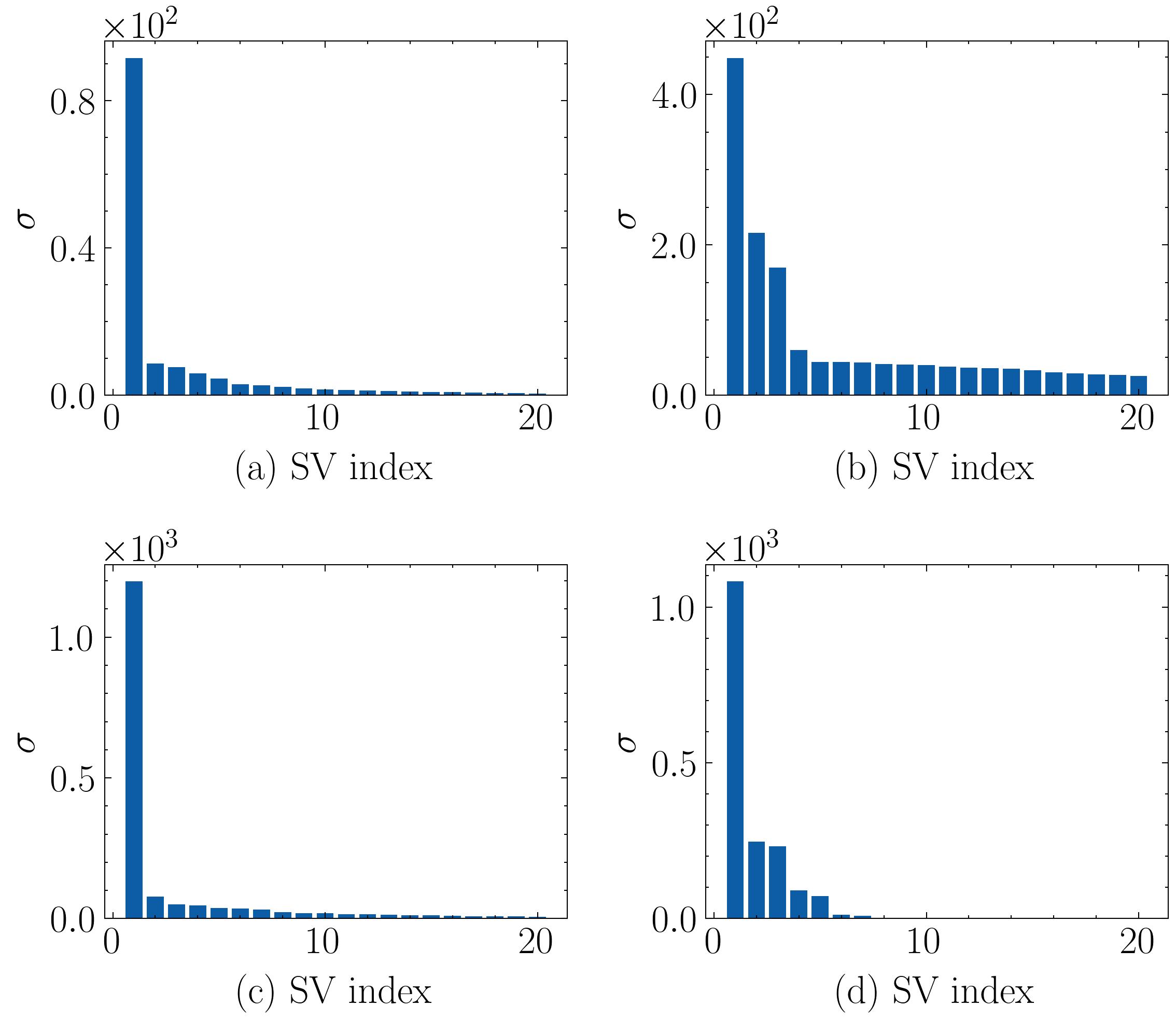}
    \caption{Singular values (SV) of $\hbQ$ obtained via $Q$-learning in four standard RL problems: (a) Pendulum, (b) Cartpole, (c) Mountain car, and (d) Acrobot. The energy of $\hbQ$ tends to concentrate on the main SVs.}
    \label{fig:example_q_learning}
\end{figure}

While analyzing the rank of the optimal $\bbQ^*$ is of great value, in RL problems the MDP variables $\bbr$, $\bbP$, and $\bbPi$  are not typically known and one must resort to TD methods to estimate the SA-VF. Although some TD algorithms, such as $Q$-learning, are guaranteed to converge to the true SA-VF, the required assumptions are not always easy to hold in practical scenarios. In particular, while strict convergence requires every state to be visited infinitely often, in many real setups some states are hardly visited. Remarkably, the SA-VF matrix $\hbQ$ estimated in those cases, while different from $\bbQ^*$, can still be used to induce a good-performing policy. Hence, a prudent question in those environments is whether matrix $\hbQ$ is approximately low-rank. To test this, we run $Q$-learning in various RL environments and analyze the singular values of  $\hbQ$. The results are reported in Fig. \ref{fig:example_q_learning}, where, once again, we observe that just a few singular values are large, providing additional motivation for the algorithms designed in this paper. It is worth pointing out that, although related, this observation is different from the assumption made in Proposition \ref{theorem} and Corollary \ref{corollary}. Indeed, the assumptions are about the error in approximations on trajectories of the value function approximation, while the experiments in Fig. \ref{fig:example_q_learning} are only for the optimal Q function.

\begin{figure}
    \centering
    \includegraphics[width=87mm]{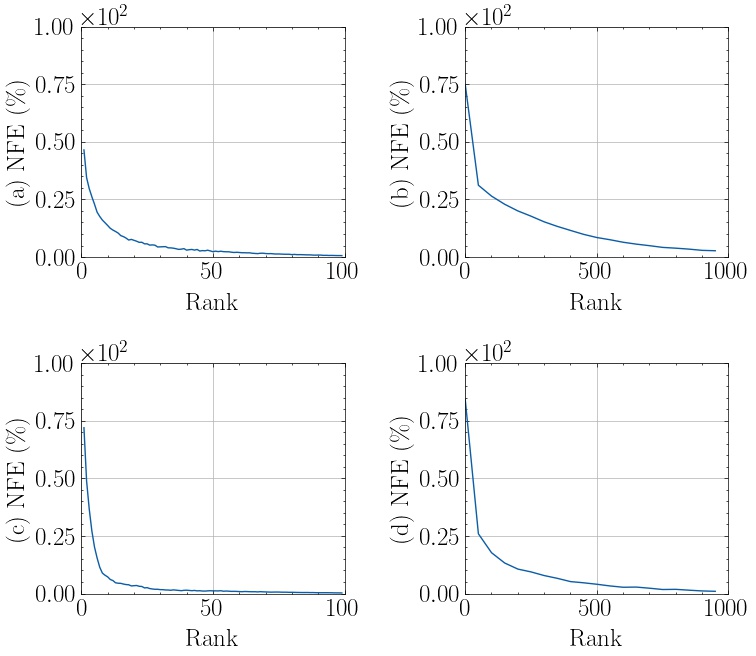}
    \caption{$\mathrm{NFE}$ between the tensor form of $\hbQ$ obtained via $Q$-learning and its low-rank PARAFAC decomposition in four standard RL problems: (a) Pendulum, (b) Cartpole, (c) Mountain car, and (d) Acrobot. We observe how the error decreases as the rank of the approximation increases.}
    \label{fig:example_q_learning_tensor}
\end{figure}

The next step is to check if the low-rank approximation is also accurate when the SA-VF is represented by a tensor. To that end, we re-arrange the $Q$-matrices obtained in the $Q$-learning experiments as $Q$-tensors. Then, we decompose the $Q$-tensor using the PARAFAC decomposition for various ranks. To measure the approximation performance we use the Normalized Frobenius Error ($\mathrm{NFE}$) between a given tensor $\tenbX$ and its approximation $\check{\tenbX}$, which is given by
\begin{equation}
    \mathrm{NFE}={||\tenbX - \check{\tenbX}||_F}\big/{||\tenbX||_F}.
    \label{eq:NFE}
\end{equation}
In this experiment, the original tensor is the one obtained via $Q$-learning and the approximation is the PARAFAC decomposition. The results are shown in Fig. \ref{fig:example_q_learning_tensor} and the patterns observed are consistent. While the error decreases as the rank increases, most of the energy is concentrated in the first factors, so that relatively low-rank approximations of the $Q$-tensor are close (in a Frobenius sense) to the original values estimated by $Q$-learning.

So far, we have shown that low-rank approximations of the SA-VF can be expected to be similar to their full-rank counterparts. However, in most cases the interest in not in the SA-VF itself, but in the policy (actions) inferred from it. Hence, a natural question is whether the policies induced by the low-rank SA-VF approximation lead to high rewards. With this question in mind, we test the policies obtained from the truncated SVD of $\hbQ$ in various environments. We measure the Normalized Cumulative Reward Error $\mathrm{NCRE}=(r - \hat{r})\big/ r$, where $r$ denotes the cumulative reward obtained using the original policy, and $\hat{r}$ the cumulative reward obtained using the policy inferred from the truncated SVD. The results, reported in Table \ref{tab:svd-error}, confirm that the reward obtained using the low-rank induced policies is indeed close to the reward obtained using the original policies. In four out of six scenarios, the $\mathrm{NCRE}$ error is below $1 \%$, providing additional support for our approach.

{
\renewcommand{\arraystretch}{1.1} 
\begin{table}[h!]
\centering
\begin{tabular}{l||l|l|l}
\hline
\textbf{Environment} & \textbf{Frozen Lake} & \textbf{Race Track} & \textbf{Rental Car}  \\ 
\hline
\textbf{NCRE (\%)} & 0.00 & 10.05 & 0.84 \\
\hline
\multicolumn{4}{c}{} \\
\hline
\textbf{Environment} & \textbf{Pendulum}  & \textbf{Cartpole}   & \textbf{Mountain Car} \\ 
\hline
\textbf{NCRE (\%)} & 0.42 & 5.75 & 0.10 \\
\hline

\end{tabular}
\vspace{3mm}
\caption{Error obtained with a truncated-SVD-based policy.}
\label{tab:svd-error}
\end{table}
}

\vspace{-5mm}

\subsection{Parameter efficiency}

Compared to traditional RL schemes, low-rank algorithms lead to similar cumulative rewards while having fewer parameters to learn. These complexity savings are particularly relevant in setups with continuous state-action spaces. In such a context, it is common to use non-parametric models with a discretized version of the state-action spaces. The finer the discretization, the larger the SA-VF matrix. Finer discretizations lead to better results (more expressive) but increase the size of the $Q$-matrix to estimate. More specifically, the number of parameters scales exponentially with the number of dimensions. The low-rank algorithms proposed in this manuscript enable finer discretizations while keeping the number of parameters of the model under control. 

To test this point, we present an experiment comparing the MLR algorithm (MLR-learning), the TLR algorithm (TLR-learning), $Q$-learning, deep $Q$-learning\footnote{DQN is arguably the state-of-the-art algorithm in value-based RL. Succinctly, DQN considers Neural Networks (NNs) to map the states into the $Q$-values of the actions, with the training of the NNs being carried out offline. Specifically, DQN stores the trajectories in an Experience Replay (ER) buffer \cite{mnih2013playing} to then sample from the buffer state-action-reward triplets to train. For the DQN simulations in this paper, we use a uniform ER buffer in all scenarios, except for the mountain car problem, where we used a Prioritized ER buffer \cite{schaul2015prioritized} so that we could deal with sparse rewards in a more efficient manner.}  (DQN) \cite{mnih2013playing}, and structured value-based RL (SV-RL) \cite{yang2019harnessing}, which is a variant of DQN that solves a (low-rank) matrix factorization problem in each time-step to estimate the TD targets. DQN and SV-RL can handle continuous state spaces, hence these setups do not require the state space to be discretized. Furthermore, the hyper-parameters of DQN and SV-RL have been adjusted to use the most compact (simplest) architectures able to solve the proposed tasks.  We tested the algorithms in 4 standard RL problems \cite{brockman2016openai}--\cite{Goddard2021}:
\begin{itemize}
    \item The pendulum problem tries to keep a pendulum pointing upwards and consists of $D_\ccalS=2$ state dimensions and $D_\ccalA=1$ action dimension. The reward function  weights the position of the pendulum and the torque developed to control it. The original implementation has been modified to increment the weight of the torque in the reward function, stressing the relevance of balancing the pendulum with the lowest possible control action, and thus making the problem harder.
    \item The cartpole problem again tries to keep a pendulum under control, but this time with the pendulum placed in a cart that moves horizontally. The problem consists of $D_\ccalS=4$ state dimensions and $D_\ccalA=1$ action dimension. The original problem has a discrete (binary) action space. We have modified it to be continuous to allow a richer variety of policies. Originally, the reward function is $+1$ for every time step that the pendulum is pointing upwards. However, this does not reflect the importance of selecting the appropriate control action in a continuous action-space setup. We have modified it to weigh the position of the pole and the magnitude of the control action. Controlling the pole by developing low horizontal forces is required to succeed in the problem.
    \item The mountain car problem, where a car tries to reach the top of a hill, and consists of $D_\ccalS=2$ state dimensions and $D_\ccalA=1$ action dimension.
    \item The Goddard-rocket problem tries to get a rocket as high as possible controlling the amount of fuel burned during the take-off. The problem consists of $D_\ccalS=3$ state dimensions and one $D_\ccalA=1$ dimension.
\end{itemize}
Our implementations use the toolkit OpenAI~Gym \cite{brockman2016openai}, \cite{Goddard2021}, with the full implementation details being provided in \cite{Rozada2023}. 

Let us denote the discretization of the continuous state space $\ccalS$ as $\ccalS^r$, and the discretization of the action space $\ccalA$ as $\ccalA^r$, where the subindex $r$ represents a resolution level. The finer the resolution level, the better the approximation of the continuous state-action spaces. However, this affects the number of parameters of the model, since it acts on the finite number of states $C_{\ccalS^r}$ and actions $C_{\ccalA^r}$. We test various resolution levels. We run each experiment $100$ times until convergence using a decaying $\epsilon$-greedy policy to ensure exploratory actions at the beginning of the training. The updates of the models take place in every time-step, right after sampling a new transition. The partition-based approach explained in Section \ref{Ss:matrix_and_tensor_stochastic} has been leveraged in the Goddard-rocket problem to enhance the expressiveness of the model by grouping two dimensions. Table \ref{tab:parameters-table} lists the median cumulative reward obtained by agents of different sizes that implement a greedy policy based on the estimated SA-VFs. As expected, $Q$-learning needs more parameters to achieve a reward comparable to that of either MLR-learning or TLR-learning. DQN and SV-RL achieve high rewards with fewer parameters than $Q$-learning too, but MLR-learning and TLR-learning obtain even higher rewards in the Goddard rocket problem, as well as in the cartpole problem, where both DQN and SV-RL fail to obtain good results. Although the original implementation of the cartpole problem is an easy task to solve, the modified version with a continuous action space and a reward that penalizes large actions is a non-trivial task, as demonstrated empirically. In the mountaincar problem, MLR-learning and TLR-learning perform similarly to DQN and SV-RL. The performance of the TLR-learning algorithm is particularly relevant since it reveals that low-rank tensor models with a small number of parameters are able to achieve rewards as good as (or even better than) those by MLR-learning, DQN-learning, and SV-RL.

\subsection{Computational complexity}

The next two sections analyze the computational load required to run the different RL schemes. In this first section, we focus on computational load per update. To that end, Table \ref{tab:parameters-table} reports the (average) time taken for a single update step across different agents. Not surprisingly, $Q$-learning emerges as the lightest option, primarily because its update step does not involve evaluating a gradient. Moreover, in $Q$-learning the number of parameters does not influence the update time. This is because the number of parameters directly corresponds to the number of SA-VFs to estimate, and each update step affects only one SA-VF. A larger state space necessitates more time-steps for convergence, yet it does not increase the burden of individual update steps.

Both MLR-learning and TLR-learning operate within a similar order of magnitude as $Q$-learning. Interestingly, scaling these models does not  impact the update time. As the size of the state-action space remains fixed, the number of parameters depends on the rank $K$. Although the update time should correlate with the rank $K$, given that the required ranks to solve the problem are small, other factors predominantly influence the update time. As both MLR-learning and TLR-learning implement an alternating minimization scheme, the update time mainly hinges on the number of dimensions to update. This clarifies why TLR-learning exhibits slower update times compared to MLR-learning, despite having fewer overall parameters to estimate.

In contrast, DQN incurs a substantially higher computational complexity cost, being approximately an order of magnitude slower than its counterparts. This slowness mainly arises from its utilization of batches of samples for computing the updates. Implementing DQN using just one sample (as in $Q$-learning, MLR-learning, and TLR-learning) leads to slower convergence, as it will be explained in the next section. Additionally, scaling up the DQN model significantly affects its update time. The computational challenge with SV-RL is even more pronounced, as it tackles a matrix factorization problem in each time-step. Consequently, the computational load per update step in SV-RL far surpasses that of all other alternatives.


\begin{table*}[h]
\begin{tabular}{cc}%
\begin{tabular}{l||llll}
\hline
\textbf{Environment}           & \textbf{Algorithm}            & \textbf{Return} & \textbf{\#Params.}  & \textbf{Time ($\mu$s)} \\ \hline
\multirow{18}{*}{\textbf{Pendulum}}     & \multirow{4}{*}{$Q$-learning}   & 89.55                      & 800              &  23.38 \\
                               &                               & 95.73                      & 1,600           &   23.08 \\
                               &                               & 96.39                      & 2,400           &   22.94  \\
                               &                               & 96.69                      & 4,000           &   23.00 \\ \cline{2-5} 
                               & \multirow{4}{*}{MLR-learning} & 93.85                      & 820             &   31.46 \\
                               &                               & 95.45                      & 1,640           &   31.25 \\
                               &                               & 95.39                      & 2,460           &   31.85 \\
                               &                               & 95.49                      & 4,100           &   31.94 \\ \cline{2-5} 
                               & \multirow{4}{*}{TLR-learning} & 93.94                      & 100             &   59.24 \\
                               &                               & 94.66                      & 200             &   59.98 \\
                               &                               & 94.61                      & 300             &   59.55 \\
                               &                               & 94.60                      & 500             &   59.40 \\ \cline{2-5} 
                               & \multirow{3}{*}{DQN}          & 97.23                      & 2,320           &   742.02 \\
                               &                               & 96.66                      & 23,020          &   954.96 \\
                               &                               & 96.69                      & 230,020         &   2,976.28 \\ \cline{2-5} 
                               & \multirow{3}{*}{SV-RL}        & 97.32                      & 2,320       &    1,531.66    \\
                               &                               & 97.34                      & 23,020       &   4,075.71    \\
                               &                               & 97.70                      & 230,020      &   6,707.60   \\ \hline
\multirow{18}{*}{\textbf{Mountain Car}} & \multirow{4}{*}{$Q$-learning}   & 74.99.                     & 400        &     20.13    \\
                               &                               & 83.17                      & 800          &     20.15  \\
                               &                               & 82.40                      & 1,200        &     20.06 \\
                               &                               & 83.78                      & 2,000        &     20.23 \\ \cline{2-5} 
                               & \multirow{4}{*}{MLR-learning} & -1000.00                   & 420          &     27.90  \\
                               &                               & 80.75                      & 840          &     27.94  \\
                               &                               & 85.75                      & 1,680        &     27.90  \\
                               &                               & 86.87                      & 2,100        &     28.86  \\ \cline{2-5} 
                               & \multirow{4}{*}{TLR-learning} & -1000.00                   & 80           &     46.14  \\
                               &                               & 82.45                      & 160          &     47.36  \\
                               &                               & 86.75                      & 240          &     48.54  \\
                               &                               & 88.02                      & 400          &     46.59  \\ \cline{2-5}
                               & \multirow{3}{*}{DQN}          & 88.70                      & 1,310        &   1,205.37    \\
                               &                               & 88.40                      & 13,010       &   1,485.48    \\
                               &                               & 89.20                      & 130,010      &   3,543.48    \\ \cline{2-5}
                               & \multirow{3}{*}{SV-RL}        & 88.16                      & 1,310      &    1,904.20     \\
                               &                               & 88.47                      & 13,010     &      2,155.85   \\
                               &                               & 88.44                      & 130,010      &     4,127.91  \\ \hline
\end{tabular}
\hfill \break

\begin{tabular}{l||llll}
\hline
\textbf{Environment}           & \textbf{Algorithm}            & \textbf{Return} & \textbf{\#Params.}     & \textbf{Time ($\mu$s)} \\ \hline
\multirow{18}{*}{\textbf{Cartpole}}     & \multirow{4}{*}{$Q$-learning}   & -44.59                     & 80,000       &   25.85     \\
                               &                               & 35.08                      & 160,000        &   25.84  \\
                               &                               & 70.24                      & 240,000        &   26.12  \\
                               &                               & 83.51                      & 400,000        &   25.65  \\ \cline{2-5} 
                               & \multirow{4}{*}{MLR-learning} & 6.91                       & 80,020         &   36.38  \\
                               &                               & 89.46                      & 160,040        &   36.27  \\
                               &                               & 89.43                      & 240,060        &   36.17  \\
                               &                               & 89.53                      & 400,100        &   36.60 \\ \cline{2-5} 
                               & \multirow{4}{*}{TLR-learning} & -86.10                     & 140            &   75.09  \\
                               &                               & 88.55                      & 280            &   75.06  \\
                               &                               & 89.59                      & 420            &   74.95  \\
                               &                               & 89.57                      & 700            &   74.59  \\ \cline{2-5} 
                               & \multirow{3}{*}{DQN}          & -94.77                     & 1,510          &   711.91  \\
                               &                               & -93.60                     & 15,010         &   934.60  \\
                               &                               & -94.99                     & 150,010        &   2,993.98  \\ \cline{2-5}
                               & \multirow{3}{*}{SV-RL}        & -94.55                      & 1,510      &    3,614.89   \\
                               &                               & -94.40                      & 15,010       &    4,086.06 \\
                               &                               & -94.13                      & 150,010       &   6,450.52   \\ \hline
                               
\multirow{18}{*}{\textbf{Goddard}}      & \multirow{4}{*}{$Q$-learning}   & 115.60                     & 16,000       &    25.08   \\
                               &                               & 116.62                     & 32,000       &    24.80   \\
                               &                               & 117.78                     & 64,000       &    24.30   \\
                               &                               & 117.99                     & 80,000       &    24.61   \\ \cline{2-5} 
                               & \multirow{4}{*}{MLR-learning} & 120.01                     & 16,020       &    35.22  \\
                               &                               & 117.05                     & 32,040       &    34.67   \\
                               &                               & 120.31                     & 48,060       &    35.04   \\
                               &                               & 125.94                     & 80,100       &    34.81  \\ \cline{2-5} 
                               & \multirow{4}{*}{TLR-learning} & 119.05                     & 860          &    64.27   \\
                               &                               & 119.17                     & 1,720        &    64.30   \\
                               &                               & 119.01                     & 2,580        &    64.30   \\
                               &                               & 119.31                     & 4,300        &    63.95   \\ \cline{2-5}
                               & \multirow{3}{*}{DQN}          & 114.15                     & 1,410        &    564.17   \\
                               &                               & 117.60                     & 14,010       &    715.00   \\
                               &                               & 117.78                     & 140,010      &    1,997.78   \\ \cline{2-5}
                               & \multirow{3}{*}{SV-RL}        & 119.84                      & 1,410        &    1,352.67   \\
                               &                               & 119.23                      & 14,010    &     2,229.98     \\
                               &                               & 119.10                      & 140,010       &   2,639.12   \\ \hline
\end{tabular}
\end{tabular}
\hfill \break
\caption{\label{tab:parameters-table}Parameters vs. Return vs. Update time.}
\end{table*}

\subsection{Speed of convergence}

MLR-learning and TLR-learning models exhibit advantages in terms of convergence speed with respect to non-parametric models. While the former updates the rows and the columns of the matrices associated with a given state-action pair, the latter only updates one entry of the $Q$-matrix at a time. Thus, low-rank SA-VF models are expected to converge faster than non-parametric models. 

\begin{figure}
    \centering
    \includegraphics[width=87mm]{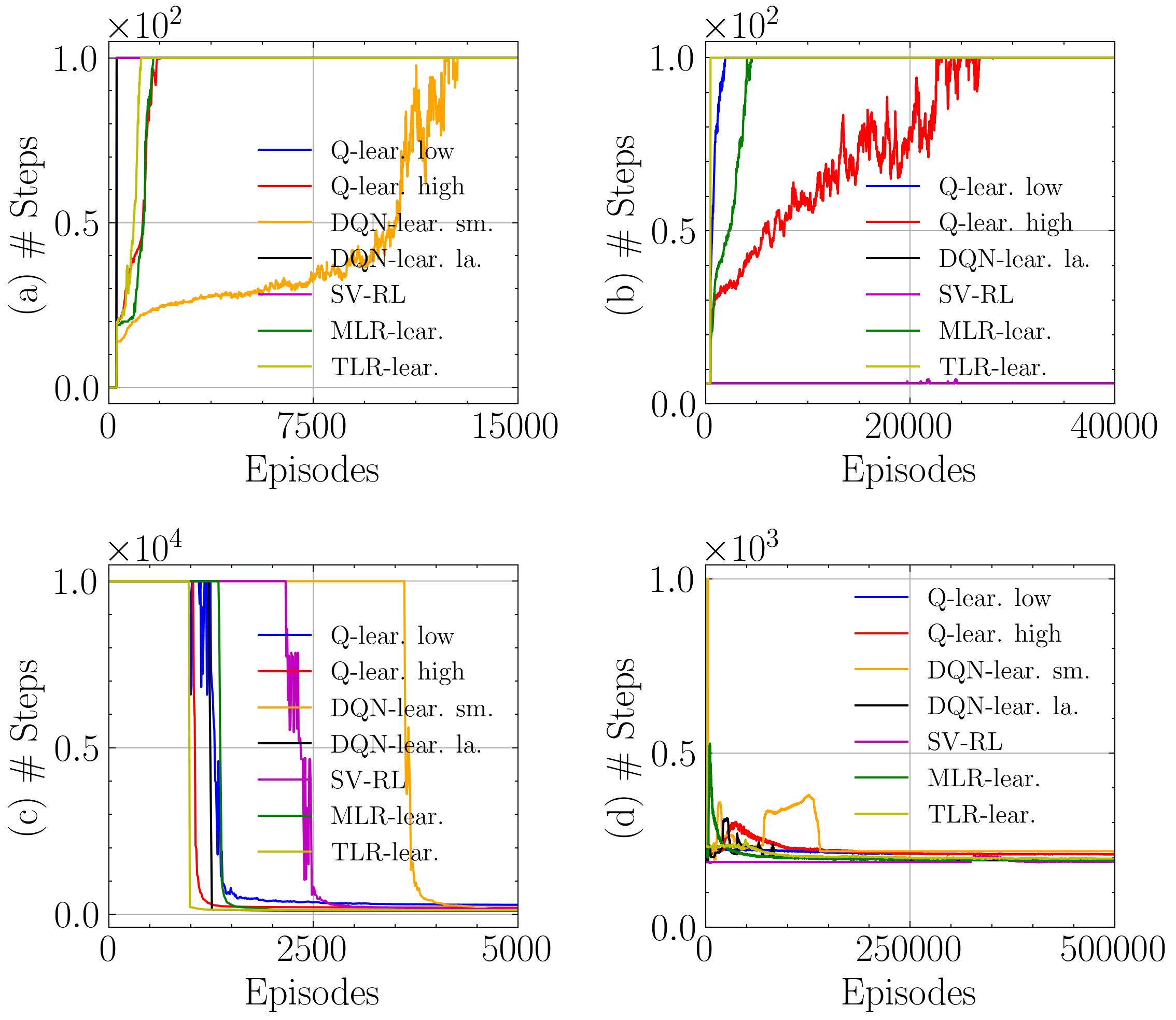}
    \caption{Median number of steps per episode in 4 standard RL problems: (a) Pendulum, (b) Cartpole, (c) Mountain car, and (d) Goddard rocket. The MLR and TLR algorithms are faster than DQN and SV-RL, especially when the batch size in DQN is large. It is also faster than $Q$-learning when the $Q$-matrix is large.}
    \label{fig:exp_speed_steps}
\end{figure}

\begin{figure}
    \centering
    \includegraphics[width=87mm]{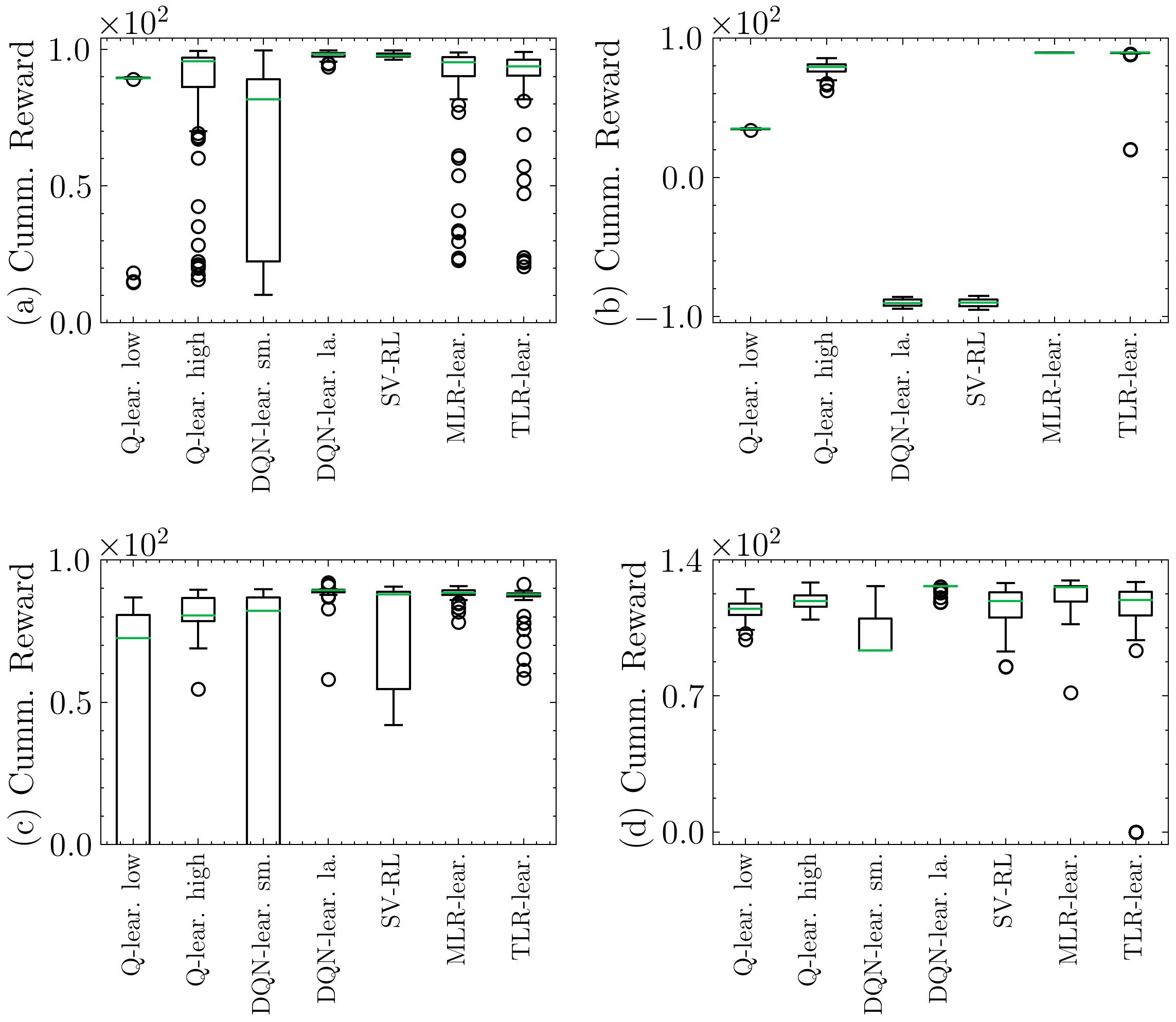}
    \caption{Distribution of the mean cumulative rewards per episode of $100$ agents in four standard RL problems: (a) Pendulum, (b) Cartpole, (c) Mountain car, and (d) Goddard rocket. On the one hand, the distributions of $Q$-learning are always shifted toward worse results. On the other hand, the dependency of DQN on the size of the ER buffer is clear.}
    \label{fig:exp_speed_reward}
\end{figure}

As in the previous section, we tested the same finer and coarser versions of $Q$-learning, DQN with a mono-sample, and a multi-sample ER scheme, and SV-RL with a multi-sample ER of the same size of the one used in DQN. We compared them to MLR-learning and TLR-learning. The agents were trained for a fixed number of episodes using a decaying $\epsilon$-greedy policy with $\epsilon$ being high at first to ensure exploratory actions. During the training, we run episodes using a greedy policy with respect to the estimated SA-VF to filter out the exploration noise. The results are reported in Fig. \ref{fig:exp_speed_steps} and Fig. \ref{fig:exp_speed_reward}. While the coarser version of $Q$-learning converges fast, it fails to select high-reward actions. Fig. \ref{fig:exp_speed_reward} shows that the distribution of rewards is shifted toward values lower than those obtained by the rest of the algorithms. Both MLR-learning and TLR-learning achieve high rewards while converging faster than the finer version of $Q$-learning. Although DQN and SV-RL achieve higher rewards in the pendulum setup, they converge slower than their counterparts in most scenarios. The dependency of DQN on the size of the ER buffer is evident. On the one hand, DQN with 1-sample ER converges noticeably slower than the rest of the algorithms, as DQN trades off learning steps and training samples. On the other hand, the cumulative reward results are not as consistent as the rest of the algorithms, showing larger interquartile ranges. A similar case applies to SV-RL. The performance of SV-RL in terms of median reward is similar to or even better than the performance of DQN in some setups. However, the variability of the results is larger, as they show larger interquartile ranges too. Moreover, as presented in the previous section, the time-per-update in SV-RL is two orders of magnitude larger than those in MLR-learning and TLR-learning. As expected, TLR-learning converges faster than MLR-learning in most scenarios. Nonetheless, when the reward function is sparse, e.g., non-zero in few state transitions, the TLR-learning algorithm exhibits some convergence issues, requiring more episodes than either MLR-learning or $Q$-learning. TLR-learning does not obtain better reward results than MLR-learning, but it converges faster and requires far fewer  parameters, as stated previously.

\subsection{High-dimensional setting}

As already explained, one of the main advantages of tensor low-rank models is that the number of parameters required to represent them scales linearly (and not exponentially) with the number of dimensions. This property makes them particularly handy when dealing with RL settings where the state (action) space is high dimensional. To further validate this, we tested our TLR algorithm in two environments: i) the highway environment \cite{highway-env}, where a vehicle is driving in a highway trying to reach a high speed but avoiding collisions, and ii) a wireless communications environment, where a user transmits packets to an access point. 

\noindent \textbf{Highway environment.} This environment has $D_\ccalS=9$ continuous state dimensions and one $D_\ccalA=1$ discrete action dimension. We transformed the continuous state space $\ccalS$ into a discrete state space $\ccalS^r$ as described in the previous section. Modeling the highway environment using $Q$-learning and MLR-learning is infeasible since even a coarse grid of $5-7$ points per dimension leads to $Q$-matrices with close to $10^8$ entries. Thus, we compared the TLR algorithm with DQN, which, as already explained, is one of the strongest value-based baselines.

\begin{figure}
    \centering
    \includegraphics[width=85mm]{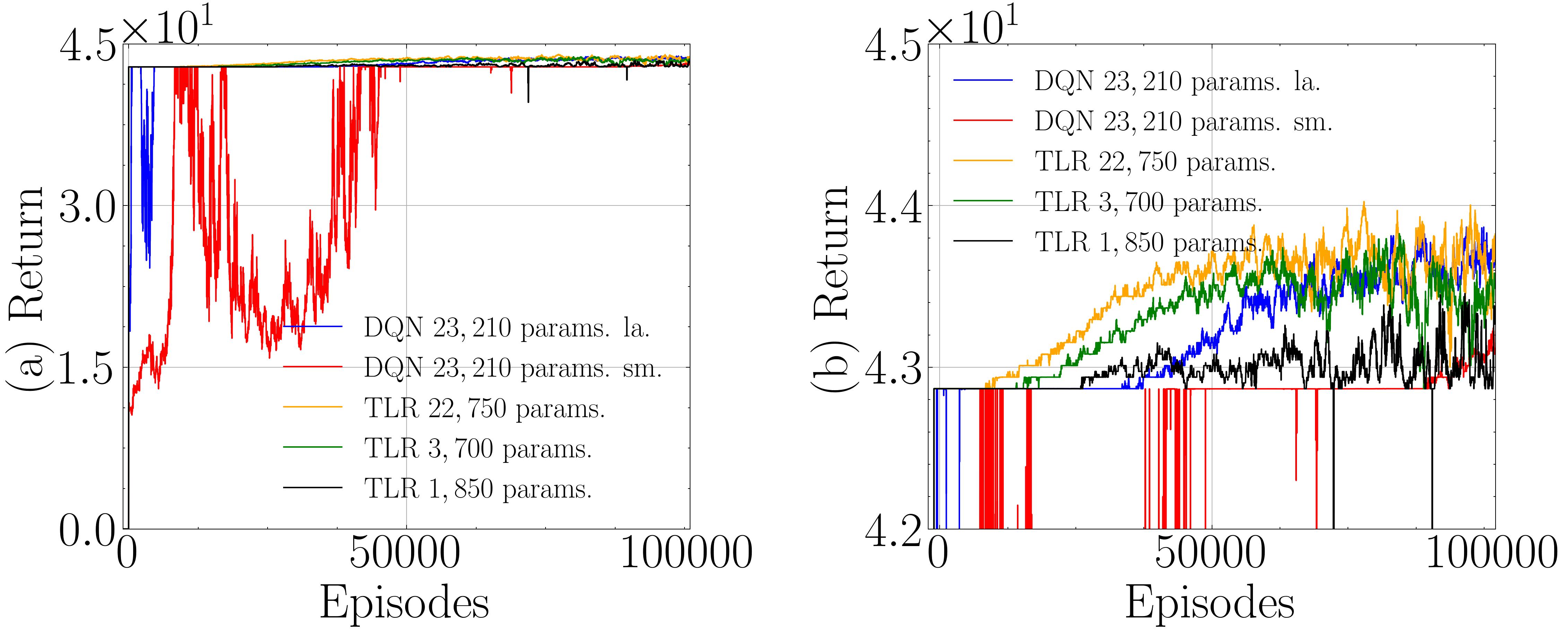}
    \caption{Median return per episode in the highway environment. The TLR algorithm has been tested with different parametrizations. The DQN algorithm has been tested with batch sizes of $1$, and $32$ respectively. The TLR converges faster, while the DQN needs more samples to achieve a good return per episode. With a small batch size, the DQN algorithm gets stuck in a local minimum.}
    \label{fig:results_highway}
    \vspace{-4mm}
\end{figure}

The results are shown in Fig. \ref{fig:results_highway}. We compared three variants of the TLR-learning algorithm (each associated with different values of the resolution level $r$ and the tensor rank $K$, affecting the total number of parameters to be learned). Then, we trained two variants of the DQN algorithm. Both variants have the same number of parameters, but the size of the batch sampled to train is different. In one case, just $1$ sample is used per training step, while in the other the size is $32$. The DQN algorithm trained with the $32$-samples batches and the two TLR algorithms with more parameters converge to a higher-return-per-episode solution. However, the TLR algorithms exhibit a faster convergence rate. An interesting point to note here is that the TLR version with the most parameters is the fastest to converge. In contrast, the DQN  trained with $1$ sample per step converges more slowly and gets stuck in a local minimum, achieving a worse return-per-episode solution. The TLR version with fewer parameters obtains a middle-point solution between the previously described algorithms.  In a nutshell, DQN needs far more samples than TLR learning to obtain a reasonable result, and even so, the convergence rate is worse.

\noindent \textbf{Wireless communications environment.} Here we simulate an opportunistic multiple-access wireless setup. We focus on a single agent, equipped with a queue and a battery, who transmits packets to an access point. There are $C=3$ orthogonal channels, and new packets arrive every $T=10$ time-slots. The user accesses opportunistically, but the channel may be occupied. The state of the system accounts for a total of $D_\ccalS=8$ dimensions: a) the fading level and the occupancy state of each of the $C=3$ channels, b) the energy in the battery, and c) the number of packets in the queue.  In each time-step, the agent selects the (discretized) power to send through each of the $C=3$ channels. Therefore, the action space has $D_\ccalA=3$ dimensions. The rate of transmission is given by Shannon's capacity formula. There is a packet loss of $80\%$ when the channel is occupied. The reward blends the battery level (positively weighted) with the remaining queue size (negatively weighted), compelling the agent to balance transmission throughput and battery. Additional details on the setup parameters can be found in \cite{Rozada2023}.

The state space $\ccalS$ is a mixture of continuous and discrete dimensions (binary occupancy states), so we discretize the continuous state dimensions leading to a discrete state space $\ccalS^r$. Similarly, we discretize the action space into $\mathcal{A}^r$. A singular challenge within this problem lies in the multidimensional nature of the action space, since the number of actions of the discrete action space $C_{\ccalA^r}$ corresponds to the Cartesian product of the dimensions of $\mathcal{A}^r$. $Q$-learning and MLR-learning are unsuitable choices due to the size of the state-action space. Once more, our comparison focuses on evaluating TLR-learning against DQN.

\begin{figure}
    \centering
    \includegraphics[width=73mm]{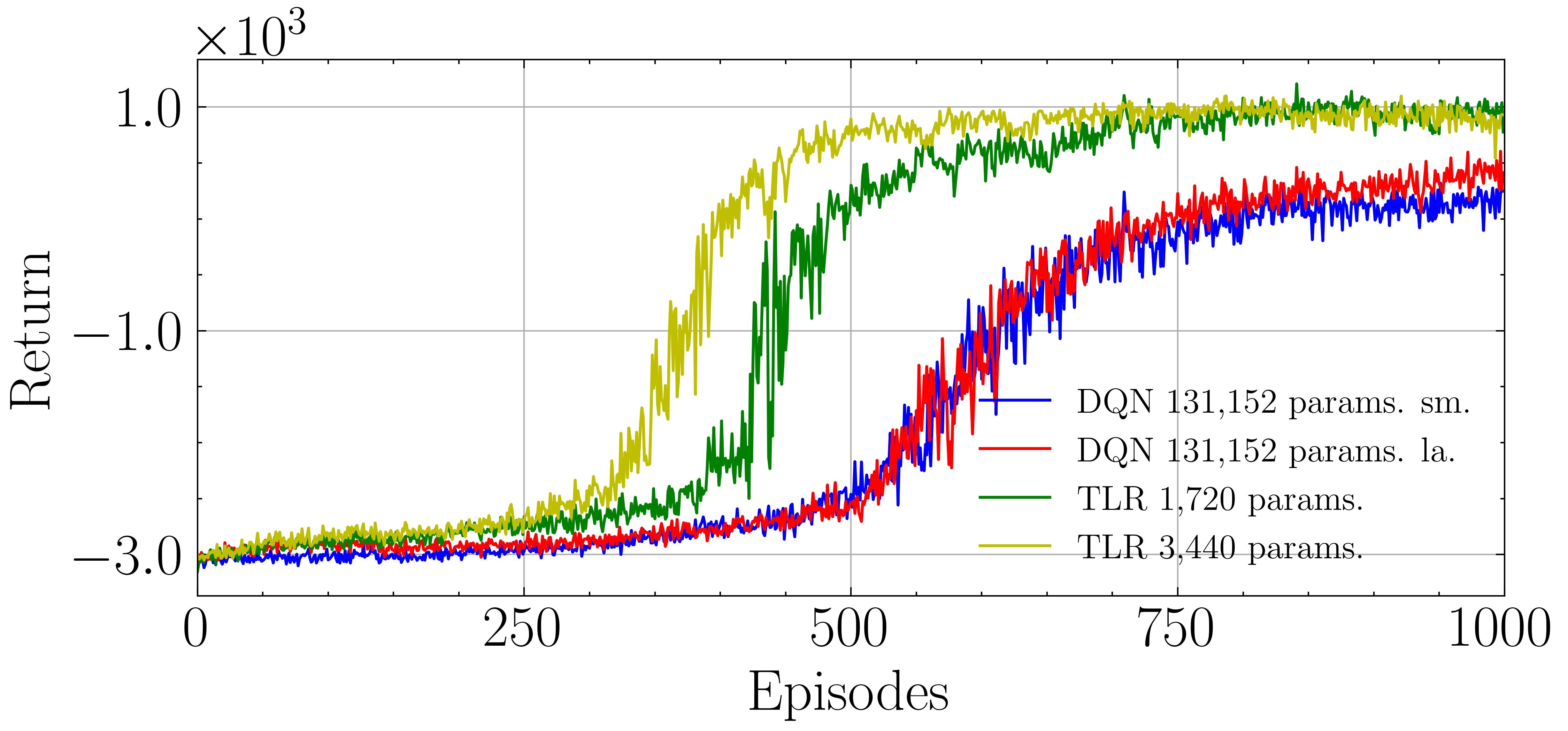}
    \caption{Median return per episode in the wireless communications setup. The TLR algorithm has been tested with different parametrizations. The DQN algorithm has been tested with batch sizes of $1$, and $16$ respectively. The TLR converges faster, while the DQN does not converge faster with more samples.}
    \label{fig:results_wireless}
\end{figure}

Fig. \ref{fig:results_wireless} validates the findings observed in the highway environment, but in a very different setup. We compared two TLR-learning variants with ranks $K=20$ and $K=40$ respectively, against two DQN variants, one with a batch size of $1$ and the other with a batch size of $16$. Notably, TLR converges faster than DQN in both cases, with the version having more parameters once again showing the fastest convergence. In contrast, DQN converges slower, independently of the batch size. These results illustrate that there exists realistic high-dimensional RL problems that can be efficiently handled by our TLR algorithm, requiring fewer resources and converging faster than standard DL-based approaches.

\section{Conclusion}

This paper introduced a range of low-rank promoting algorithms to estimate the VF associated with a dynamic program in a stochastic, online, model-free fashion. We first kept the classical state-action matrix representation of the VF function and developed matrix low-rank stochastic algorithms for its online estimation. Since most practical problems deal with high-dimensional settings where both states and actions are vectors formed by multiple scalar dimensions, we then proposed a tensor representation for the VF function. The subsequent natural step was the development of low-rank tensor-decomposition methods that generalize those for the matrix case. The savings in terms of number of parameters to be estimated was discussed and the proposed algorithms were tested in a range of reinforcement learning scenarios, assessing the performance (reward and speed of convergence) for multiple configurations and comparing it with that of existing alternatives. The results are promising, opening the door to the adoption of stochastic non-parametric VF approximation (typically confined to low-dimensional RL problems) in a wider set of scenarios.

\bibliographystyle{IEEEtran}
\bibliography{journal}



\end{document}